\documentclass{article}

\PassOptionsToPackage{numbers}{natbib}
\usepackage[preprint]{neurips_2026}
\usepackage[pagebackref=true]{hyperref}
\hypersetup{
  pdftitle={Complete or Sparse: A Tale of Two Identifiabilities},
  pdfauthor={Junze Zhou, David Klindt}
}
\renewcommand{\backref}[1]{} 
\renewcommand{\backrefalt}[4]{%
  \ifcase #1 
    (Not cited.)%
  \or 
    (Cited on page #4.)%
  \else 
    (Cited on pages #4.)%
  \fi%
}

\usepackage[utf8]{inputenc} 
\usepackage[T1]{fontenc}    
\usepackage{url}            
\usepackage{booktabs}       
\usepackage{nicefrac}       
\usepackage{microtype}      
\usepackage{xcolor}         
\usepackage{wrapfig}
\usepackage{tcolorbox}
\usepackage{amsmath}
\usepackage{amsthm}
\usepackage{amsfonts}
\usepackage{amssymb}
\usepackage[inline]{enumitem}   
\usepackage[capitalise,noabbrev]{cleveref} 

\usepackage{graphicx}
\usepackage{tikz}
\usetikzlibrary{arrows.meta, positioning, shapes.geometric, fit, backgrounds, calc}
\usepackage{placeins}
\usepackage{bm}
\usepackage{bbm}

\newtheorem{theorem}{Theorem}[section]
\newtheorem*{theorem*}{Theorem}
\newtheorem{lemma}[theorem]{Lemma}

\newtheorem*{corollary*}{Corollary}
\newtheorem{remark}{Remark}

\newtheorem{assumption}{Assumption}[section]  

\newcommand{\supp}{\operatorname{supp}}

\newcommand{\T}{^{\mathsf{T}}}

\newcommand{\KL}{\mathrm{KL}}            
\newcommand{\CE}{\mathrm{CE}}            

\DeclareMathOperator{\diag}{diag}
\DeclareMathOperator{\rank}{rank}
\DeclareMathOperator{\Tr}{Tr}

\DeclareMathOperator{\Span}{Span}
\DeclareMathOperator{\Row}{Row}
\DeclareMathOperator{\Ker}{Ker}
\newcommand{\one}{\mathbf{1}}            
\DeclareMathOperator{\Image}{Im}         

\newcommand{\E}{\mathbb{E}}              
\newcommand{\R}{\mathbb{R}}              

\DeclareMathOperator{\softmax}{softmax}

\title{Complete or Sparse: A Tale of Two Identifiabilities}

\author{
  Junze Zhou\\
  Cold Spring Harbor Laboratory\\
  \texttt{juzhou@cshl.edu}
  \And
  David Klindt\\
  Cold Spring Harbor Laboratory\\
  \texttt{klindt@cshl.edu} 
}

\begin{document}

\maketitle

\begin{abstract}

Understanding when neural networks learn interpretable representations is central to mechanistic interpretability. We study this through identifiability: when can supervised learning recover the true latent variables underlying the data? We introduce two \textit{data generating processes} distinguished by their \textit{causal direction} where either labels cause latents (\textit{generative}) or latents cause labels (\textit{discriminative}). When the representation has at least as many dimensions as the latent space, we prove that the optimal encoders are \textit{linear} in the latents. Conversely, when dimensions are insufficient, identifiability generically fails and encoders become \textit{nonlinear}. We then show that additional constraints yield recovery of individual latent variables, not just linear mixtures. In the generative setting, non-negative activations and energy efficiency force each neuron to encode exactly one latent, providing a precise account of selective \textit{grandmother cell}-like responses. In the discriminative setting, when latents are sparse and the true measurement matrix satisfies the spark condition, the learned encoder inherits \textit{compressed sensing} structure, enabling sparse dictionary learning to perfectly recover individual latents. This constitutes, to our knowledge, the first identifiability result for undercomplete nonlinear independent component analysis. Our framework unifies nonlinear ICA, compressed sensing, and neural network interpretability, delineating exactly when representations transition from nonlinear distortions to linear representations and, under additional constraints or steps, to clean, axis-aligned codes. Theoretical predictions are validated on synthetic benchmarks and on activations drawn from vision models, large language models, and macaque inferotemporal cortex.

\end{abstract}

\section{Introduction}\label{sec:intro}
The discovery of ``grandmother cells'' \citep{gross2002genealogy}, neurons that selectively respond to highly specific concepts such as the ``Jennifer Aniston neuron'' in human medial temporal lobe \citep{quiroga2005invariant}, has challenged traditional theories of neural coding \citep{thorpe_local_1989}. Similar observations have emerged in modern vision \citep{bau2020understanding} and large language models, exemplified by neurons selectively responding to concepts such as certain despots \citep{goh2021multimodal}. Specific concepts, encoded by individual neurons, have even been found across models \citep{dravid2023rosetta} and brains \citep{khosla_privileged_2024}. The existence of these highly interpretable individual neurons is surprising since dedicated single-neuron representations are generally assumed to be computationally inefficient and fragile \citep{rigotti2013importance}. 

Perhaps more surprising still is the presence of interpretable concepts (\textit{latent variables}) within such nonlinear systems which could, in principle, reformat information in highly unintelligible ways \citep{hyvarinen1999nonlinear}. Recent theoretical advances in identifiability theory, particularly nonlinear Independent Component Analysis (ICA), suggest that neural populations optimally recover the underlying latent variables of the data up to an arbitrary linear transformation \citep{hyvarinen1999nonlinear,hyvarinen2019nonlinear}. With such an arbitrary linear transformation, individual neurons are almost certainly responsive to linear combinations of latent variables. This means that they exhibit what is called \textit{mixed selectivity} \citep{rigotti2013importance} in neuroscience, or, \textit{polysemanticity} \citep{olah_building_2018} in mechanistic interpretability of AI research.
This stands in tension to the empirical observations of clean selective (\textit{monosemantic}), interpretable neurons \citep{quiroga2005invariant,khosla_privileged_2024,goh2021multimodal,bau2020understanding,higgins_unsupervised_2021}.

To resolve this contradiction, we introduce a sparse generative model of latent variables that better aligns theoretical predictions with empirical data. This model captures the intuitive notion that natural stimuli are sparse and conceptually distinct.
Specifically, exponential priors naturally emerge in sensory coding \cite{foldiak1990forming,olshausen1996emergence}.
This offers a principled explanation for the success of sparse autoencoders in capturing interpretable neural representations \citep{templeton_scaling_2024,bricken_towards_2023,klindt_identifying_2023}. Combined with biologically plausible assumptions such as non-negative neural activations, we prove mathematically and demonstrate empirically that the statistical optimum of neural coding under this model naturally yields individually interpretable neurons aligned directly with latent variables.

\paragraph{Contributions.} \textbf{1.} We propose a sparse Data Generating Process (DGP) (sec.~\ref{sec:dgp}) that addresses key limitations of existing nonlinear independent component models. \textbf{2.} We derive identifiability results for both the overcomplete ($m > n'$) and full-capacity ($m = n'$) settings, proving that optimal encoders recover the latent variables up to a linear transformation (\cref{thm:linear_id}). Conversely, we prove that in the undercomplete setting ($m < n'$), the global optimum of the cross-entropy loss is generically nonlinear—linear encoders are provably suboptimal (\cref{thm:nonlinear}). \textbf{3.} Integrating identifiability theory with non-negative and energy constraints, we prove that optimal solutions correspond to axis-aligned, monosemantic neurons (\cref{thm:axis}). \textbf{4.} We analyze the undercomplete setting where the number of neurons is insufficient, and show that it leads to distributed neural representations that support compressed sensing (\cref{thm:cs}), providing a theoretical explanation for the empirical success of sparse autoencoders. 
\textbf{5.} We validate our theoretical predictions empirically, bridging insights from neuroscience, machine learning, and interpretability research.

\section{Related Work}


\paragraph{Nonlinear ICA and Neuroscience.}
Recent advances in nonlinear ICA have established conditions under which latent variables can be identified, up to linear transformations, from observed data \citep{hyvarinen2016tcl, hyvarinen2019nonlinear, zimmermann_contrastive_2021, reizinger2025empirically}. These identifiability theories generally predict neural representations exhibiting \textit{mixed selectivity} \citep{barak_sparseness_2013, rigotti2013importance}, thereby posing challenges for the direct interpretation of single-neuron responses \citep{bau2020understanding, quiroga2005invariant, goh2021multimodal}.
By contrast, nonlinear ICA based on temporal autocorrelations has given stronger identifiability results up to permutation equivalences \citep{hyvarinen2017pcl, klindt_towards_2021}. Consistent with a one-to-one correspondence, disentangled representations have successfully matched neurons in macaque inferior temporal cortex \citep{higgins_unsupervised_2021}, linking theoretical identifiability with biological observations.

\paragraph{Sparse Coding and Sparse Autoencoders (SAEs).}
Sparse coding and non-negative matrix factorization \citep{olshausen1996emergence, lee1999learning} have long provided frameworks to explain the emergence of selective and interpretable neural representations. However, theoretical extensions beyond linear mixing remain limited. Notably, the uniqueness of sparse solutions, highlighted by compressed sensing theory \citep{donoho_compressed_2006}, has been extended to sparse coding scenarios \citep{hillar2015can}. Additionally, biological plausibility has been reinforced by imposing non-negativity constraints \citep{pehlevan2017blind, lipshutz2022biologically, whittington2022disentanglement}.
Recent empirical studies utilizing SAEs have demonstrated that sparse decompositions applied to neural representations yield interpretable sparse codes \citep{fyshe_interpretable_2014, yun_transformer_2021, bricken_towards_2023, klindt_identifying_2023, templeton_scaling_2024, o2024compute, simon2024interplm, klindt2025superposition}. These findings underscore the potential of sparsity as a crucial factor driving interpretability in neural representations, providing practical motivation for the theoretical considerations presented here.

\paragraph{Integration and Theoretical Contribution.}
Our work synthesizes these perspectives, explicitly bridging sparse coding frameworks with nonlinear identifiability theory. By integrating the uniqueness guarantees of non-negative coding into identifiability theory, we resolve conceptual tensions between sparse selectivity and theoretical optimality \citep{barlow_single_1972, thorpe_local_1989}, thereby offering a rigorous theoretical foundation for interpreting neural representations.

\section{Data Generating Processes}\label{sec:dgp}


We assume \textit{latent variables} \(z \in \mathcal{Z}\) that are \(n\)-dimensional, and class labels \(y \in \mathcal{Y}\) over \(|\mathcal{Y}|\) classes.
The difference between our two DGPs will be the causal direction between these: either \textit{generative} \(Y \to Z\) or \textit{discriminative} \(Z \to Y\).
The latent variables are transformed into \textit{inputs} \(x = g(z)\) by an invertible nonlinear function \(g: \mathcal{Z} \to \mathcal{X}\).
Inputs are transformed into \textit{activations} by a learned nonlinear function \(f: \mathcal{X} \to \mathcal{H}\), where the activation space \(\mathcal{H}\) is \(m\)-dimensional.
The composite function \(h: \mathcal{Z} \to \mathcal{H}\) is defined as \(h = f \circ g\).
We assume \(h\) is continuous.
Crucially, \(g\) is fixed while \(f\) must be learned.

The learning task in both settings is \textit{classification}.
We assume the true relationship between latents and labels is captured by a log-linear model:
\begin{equation}\label{eq:true_posterior}
    p(y|z) = \softmax(W^* z)
\end{equation}
for some matrix \(W^* \in \mathbb{R}^{|\mathcal{Y}| \times n}\), where \({}^*\) denotes quantities from the true data generating process. Softmax logits are invariant to a common row shift, so throughout we assume that \(W^*\) is row-centered. See \cref{lem:centering} for more details.
The \textit{effective dimensionality} \(n' = \rank(W^*)\) will play a central role in our analysis.

The learned model predicts class probabilities via a linear readout \(U \in \mathbb{R}^{|\mathcal{Y}| \times m}\) applied to the activations:
\begin{equation}\label{eq:learned_model}
    q(y|z) = \softmax(U h(z)).
\end{equation}
Training minimizes the cross-entropy loss:
\begin{equation}\label{eq:ce_loss}
    \mathcal{L}(h,U)
    = \E_Z\!\left[\CE\bigl(p(\cdot\mid Z),q(\cdot\mid Z)\bigr)\right].
\end{equation}
See complete assumption list in Supplement~\ref{app:assum}.
We now describe the two causal settings in detail.

\subsection{Generative Setting: From Labels to Latents \(Y \to Z\)}\label{sec:dgp_gen}

In this setting, we draw a label first, then a latent conditioned on that label.
This is akin to the causal process underlying MNIST: a person thinks of a digit, then draws it \citep{scholkopf2012causal}.
Prior work has assumed hyperspherical latents \citep{zimmermann_contrastive_2021,reizinger2025empirically,rusak2024infonce}, but this is incompatible with sparsity.
We propose a different model: non-negative latents with varying norms.

Specifically, we set \(\mathcal{Z} = \mathbb{R}_+^n\) and associate each class \(y\) with rate parameters \(\lambda_y \in \mathbb{R}_+^n\), giving factorized exponential conditionals:
\begin{equation}\label{eq:gen_conditional}
    p(z|y) = \prod_{i=1}^n \lambda_i(y) \exp\bigl(-\lambda_i(y) z_i\bigr) = \exp(\langle w^*_y, z \rangle) \prod_{i=1}^n \lambda_i(y)
\end{equation}
where \(w^*_y = -\lambda_y < 0\).
The conditional entropy is
\begin{equation}
    H(p(z|y)) = \sum_{i=1}^n \bigl[1 - \log \lambda_i(y)\bigr] = n - \sum_{i=1}^n \log \lambda_i(y).
\end{equation}
We assume constant conditional entropy across classes:
\begin{equation}\label{eq:const_entropy}
    \prod_{i=1}^n \lambda_i(y) = \prod_{i=1}^n \lambda_i(\tilde{y}), \quad \forall\, y, \tilde{y} \in \mathcal{Y}.
\end{equation}
With a uniform prior \(p(y) = 1/|\mathcal{Y}|\), Bayes' rule gives the posterior in the form of \eqref{eq:true_posterior}, where \(W^* \in \mathbb{R}^{|\mathcal{Y}| \times n}\) has rows \((w^*_y)^\top\).
We assume \(\rank(W^*) = n\), requiring \(|\mathcal{Y}| > n\).
Thus, in the generative setting, the effective dimensionality equals the full latent dimension: \(n' = n\).

\paragraph{Additional assumptions for axis-alignment.}
To obtain the stronger result that neurons align with individual latent dimensions (Section~\ref{sec:theory_axis}), we restrict to the exact-capacity regime
$m=n=n'$ and additionally assume:
(i) activations are non-negative, \(\mathcal{H} = \mathbb{R}_+^m\) (e.g., real neurons or ReLU networks);
(ii) a fixed within-class population variance (e.g., variance normalization); and 
(iii) an energy constraint favors small activations (e.g., via energy optimization or weight decay during training).

\subsection{Discriminative Setting: From Latents to Labels \(Z \to Y\)}\label{sec:dgp_dis}

In this setting, we draw a latent first from some prior \(p(z)\), and then draw a label conditioned on that latent.
This models the common image labeling problem—an image, containing some (imperfect) measurement of latent features, is presented to a human observer who labels it \citep{scholkopf2012causal}.

Here we allow more flexibility: \(p(z)\) may be any distribution over \(\mathcal{Z} = \mathbb{R}^n\).
The conditional over labels is log-linear as in \eqref{eq:true_posterior}: \(p(y|z) = \softmax(W^* z)_y\).

The key assumption is that \(W^*\) factorizes as
\begin{equation}\label{eq:dis_factorization}
    W^* = U^* A^*
\end{equation}
where \(A^* \in \mathbb{R}^{n' \times n}\) is a \textit{measurement matrix} and \(U^* \in \mathbb{R}^{|\mathcal{Y}| \times n'}\) is a \textit{readout matrix}.
This means the label depends on \(z\) only through the compressed measurement \(A^* z \in \mathbb{R}^{n'}\).
When \(n' < n\), not all latent dimensions are directly relevant for classification—a key difference from the generative setting.

\paragraph{Additional assumptions for compressed sensing recovery.}
To obtain the stronger result that latents can be fully recovered up to permutation (Section~\ref{sec:theory_cs}), we additionally assume:
(i) latents are \textit{sparse}, \(\|z\|_0 \leq k\) for all \(z \in \mathrm{supp}(p_Z)\); and
(ii) \(A^*\) satisfies the \textit{spark}-2$k$ condition: no nonzero \(2k\)-sparse vector lies in \(\Ker(A^*)\).
This ensures all \(k\)-sparse latents remain distinguishable after projection, making compressed sensing recovery possible \citep{rauhut2010compressive}.

\section{Theory}\label{sec:theory}
A central question is why neural representations sometimes align with individual latent concepts (\textit{monosemanticity}) and other times represent linear combinations of them (\textit{polysemanticity}).
To answer this, we must first address a fundamental prerequisite: \textit{linearity}. Before a system can be axis-aligned, it must first be linearly identifiable—that is, the encoder $f$ must successfully invert the nonlinear mixing $g$ up to a linear transformation.
We show that the answer depends critically on the data generating process and the capacity of the neural representation.
First, we establish when linear identifiability holds or fails (shared by both settings).
Then, we show how to go beyond linear mixing to recover the exact latent variables up to permutations—via two distinct paths depending on the setting.
For readers who prefer intuition over algebra, every step ends with a box that restates the key point.
For a complete list of assumptions, see Supplement~\ref{app:assum}.

\subsection{Linear Identifiability}\label{sec:theory_linear}
With supervised training, let encoder $h=f\!\circ g$ and linear read-out $U$ minimize the cross-entropy
\begin{equation}\label{eq:pop_CE}
  \CE(h,U)\;:=\;\E_Z\bigl[\CE\bigl(p(\cdot \mid Z),\softmax(Uh(Z))\bigr)\bigr].
\end{equation}
Recall that the true posterior is $p(y|z) = \softmax(W^* z)$ where $n' = \rank(W^*)$ is the effective latent dimension.\footnote{By the centering lemma (Supplement~\ref{lem:centering}), we assume without loss of generality that $W^*$ is row-centered, so rank equals the dimension of the row space.}
Building on prior theory \citep{hyvarinen2016tcl,zimmermann_contrastive_2021,reizinger2024cross}, we show that when the model has sufficient capacity, the optimal encoder must be linear.

\begin{theorem}[Linear identifiability]\label{thm:linear_id}
Let \cref{ass:general} hold with $m \geq n' = \rank(W^*)$. Every global minimizer $(U, h)$ of the cross-entropy satisfies
\[
    h(z) = Az + \nu(z), \qquad UA=W^*
\]
where $A \in \mathbb{R}^{m \times n}$ has $\rank(A) = n'$, and $\nu(z) \in \mathcal{N}(U)$ for all $z$.
\end{theorem}
\noindent Proof in Supplement~\ref{proof:linear_id}.

The term $\nu(z)$ captures the degrees of freedom in the $m - n'$ dimensions orthogonal to the readout. These dimensions do not affect the classification loss and may be nonlinear or arbitrary. The classification-relevant part $Az$ is always linear.

\begin{tcolorbox}[colback=blue!10,colframe=blue!50]
\textbf{TL;DR:} Although the network \textit{could} apply complicated nonlinear transformations, the best it can do for supervised classification is recover the latent variables up to linear mixing. Extra dimensions beyond $n'$ are unconstrained by the loss.
\end{tcolorbox}

\subsection{Non-identifiability under Capacity Constraints}\label{sec:theory_nonlinear}
What happens when there are not enough neurons? If $m < n'$, the model cannot achieve zero KL divergence, and the optimal encoder is forced to be nonlinear.

\begin{theorem}[Non-identifiability]\label{thm:nonlinear}
Let \cref{ass:general} hold with $m < n' = \rank(W^*)$. Assume that $\supp(p_Z)$ has nonempty interior $\Omega \subseteq \R^n$, that $p_Z$ is continuous and strictly positive on $\Omega$, and that for every candidate $A \in \R^{m \times n}$ under consideration, $\mathcal{N}(A) \cap \Omega \neq \emptyset$. Then the global minimizer of the cross-entropy loss is generically nonlinear.
\end{theorem}
\noindent Proof in Supplement~\ref{proof:nonlinear}.

\begin{tcolorbox}[colback=blue!10,colframe=blue!50]
\textbf{TL;DR:} With fewer neurons than the effective latent dimension, it is not possible to recover latents up to a linear transformation. The encoder must distort the geometry.
\end{tcolorbox}

\subsection{Beyond Linear: Axis Alignment and Sparse Recovery}\label{sec:theory_permutation}
Linear identifiability recovers $z$ up to an arbitrary linear transformation $A$. Can we do better? Under additional constraints, we can recover $z$ up to permutation and scaling—meaning individual neurons correspond to individual latent variables. We present two paths to this stronger result.

\subsubsection{Path A: Axis-Aligned Neurons (Generative Setting)}\label{sec:theory_axis}

In the generative setting with $m = n = n'$, we have linear identifiability. If we additionally impose non-negative activations (e.g., ReLU or biological spikes) and an energy constraint (e.g., weight decay or efficient coding), the encoder is a scaled permutation.


\begin{theorem}[Axis-aligned neurons]\label{thm:axis}
Following \cref{ass:general} and \cref{ass:gen}, in the generative setting with $m=n=n'=\operatorname{rank}(W^*)$, let $(U, h)$ be any minimizer of $\mathcal R(h) = \E_z\|h(Z)\|_2^2$ among the cross-entropy global minimizers. Then
$$
h(z)=DP_\sigma z,
$$
where $P_\sigma$ is a permutation matrix and $D=\operatorname{diag}(\beta_1,\ldots,\beta_n)$ with $\beta_j>0$. That is, the representation is identifiable up to permutation and component-wise positive scaling.
\end{theorem}
\noindent Proof in Supplement~\ref{proof:axis}.

\begin{tcolorbox}[colback=blue!10,colframe=blue!50]
\textbf{TL;DR:} With non-negative activations and energy constraints under exact capacity, each neuron encodes exactly one latent variable. This provides a possible theoretical explanation for ``grandmother cells.''
\end{tcolorbox}

\subsubsection{Path B: Compressed Sensing Recovery (Discriminative Setting)}\label{sec:theory_cs}
In the discriminative setting with $m \geq n'$, the encoder learns $h(z) = Az + \nu(z)$. What can we say about the structure of $A$? If the latents are sparse and the true projection $A^*$ satisfies the spark condition, then $A$ inherits this structure, enabling sparse recovery of $z$.

\begin{theorem}[Compressed sensing recovery]\label{thm:cs}
Following \cref{ass:general} and \cref{ass:dis}, in the discriminative setting with $W^* = U^* A^*$, assume $m\geq n'$, latents are $k$-sparse, and $\operatorname{spark}(A^*)>2k$. Let $P_U$ denote the
orthogonal projector onto $\Row(U)$ and define the classifier-visible
representation $\widetilde h(z):=P_Uh(z)=Az$. Then the learned encoder satisfies $\mathcal{N}(A) = \mathcal{N}(A^*)$. Consequently,
$\operatorname{spark}(A)=\operatorname{spark}(A^*)>2k$, and $z$ can be recovered from the unique $k$-sparse solution of
\[
    Az=\widetilde h(z).
\]
\end{theorem}
\noindent Proof in Supplement~\ref{proof:cs}.

\begin{tcolorbox}[colback=blue!10,colframe=blue!50]
\textbf{TL;DR:} With sparse latents, networks learn the best possible projector. SAEs may then recover all latent variables—explaining their empirical success in mechanistic interpretability.
\end{tcolorbox}

\subsection{Summary}\label{sec:theory_summary}
\begin{center}
\begin{tabular}{lcc}
\toprule
& \textbf{Generative} ($n' = n$) & \textbf{Discriminative} ($n' \leq n$) \\
\midrule
$m \geq n'$ & Linear identifiability & Linear identifiability \\
$m < n'$ & Nonlinear (non-identifiable) & Nonlinear (non-identifiable) \\
\midrule
\textbf{Path to permutation} & Non-negativity + energy & Sparsity + spark \\
\textbf{Result} & Axis-aligned neurons & Compressed sensing recovery \\
\textbf{Interpretation} & Grandmother cells & Sparse autoencoders \\
\bottomrule
\end{tabular}
\end{center}

Experiments in \cref{sec:experiments} corroborate these predictions.

\section{Experiments}\label{sec:experiments}

\subsection{Axis-aligned Neurons}\label{sec:exp:main}

\begin{figure}[tb]
    \centering
    \includegraphics[width=0.9\linewidth]{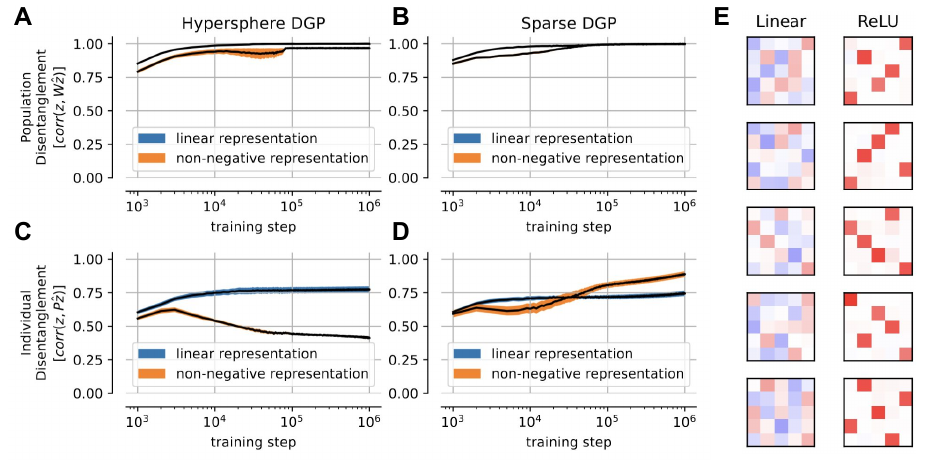}
    \caption{
    \textbf{Linear and axis-aligned latent recovery at exact capacity.}
    \textbf{A)} Hyperspherical DGP \citep{zimmermann_contrastive_2021,reizinger2025empirically}, training a neural network with linear representation (i.e., no final activation function, blue) and a non-negative representation (ReLU, orange); Disentanglement between true latent variables (\(z\)) and learned neural representations (\(\hat{z}=h(z)\)), measured up to arbitrary linear transformations (\(W\)).
    \textbf{B)} Our proposed sparse, non-negative DGP, same architectures as on the left (with/without activation function), linear disentanglement reported.
    \textbf{C)} Same as A, but disentanglement is measured only up to permutations (\(P\)), measuring axis-aligned latent variable recovery.
    \textbf{D)} Same as B but individual, axis-aligned disentanglement measured. 
    \textbf{E)} Sample visualizations of five optimization runs, learned linear representation \(h(z)=Az\) showing a clear permutation only for the non-negative case.
    Shaded areas in all plots indicate standard error of the mean over 10 random seeds.
    }
    \label{fig:lin_rep_relu}
\end{figure}

We begin with the \emph{full–capacity} setting ($m=n'$), in which our theory predicts that every neuron should align with a single latent dimension when activations are constrained to be non-negative.
We compare two DGPs: a) The hyperspherical latent model used in prior work \citep{zimmermann_contrastive_2021,rusak2024infonce,reizinger2024cross}; and b) Our proposed sparse, non-negative exponential latent model (Section~\ref{sec:dgp}). For both DGPs we train identical encoders that differ only in the final activation: a \emph{linear} head (no non-linearity) versus a \emph{non-negative} ReLU activation function.
Disentanglement is assessed with two metrics. \emph{Linear alignment} measures recovery of the latent space up to an arbitrary linear transform, whereas the stricter \emph{axis alignment} metric evaluates recovery up to permutation and element-wise scaling using the mean correlation coefficient (MCC) \citep{hyvarinen2016tcl}. Each experiment is repeated with ten random seeds.
All configurations reach near-perfect \emph{linear} alignment (Fig.~\ref{fig:lin_rep_relu}, top), showing that the encoder behaves approximately linearly across settings---confirming \cref{thm:linear_id}. However, axis alignment tells a different story (bottom panels): the combination of a sparse DGP and a non-negative output layer converges (albeit slowly) to an axis-aligned neural representation, whereas both the hyperspherical DGP and the linear output retain mixed-selectivity---confirming \cref{thm:axis}. Although convergence is slower for the non-negative encoder, we show in \cref{app:opt_dynamics} that weight decay and 
alternative activation functions substantially accelerate axis alignment.

\subsection{Non-identifiability in the undercomplete regime}\label{sec:exp:nonidentifiability}

\begin{figure}[tb]
    \centering
    \includegraphics[width=0.8\linewidth]{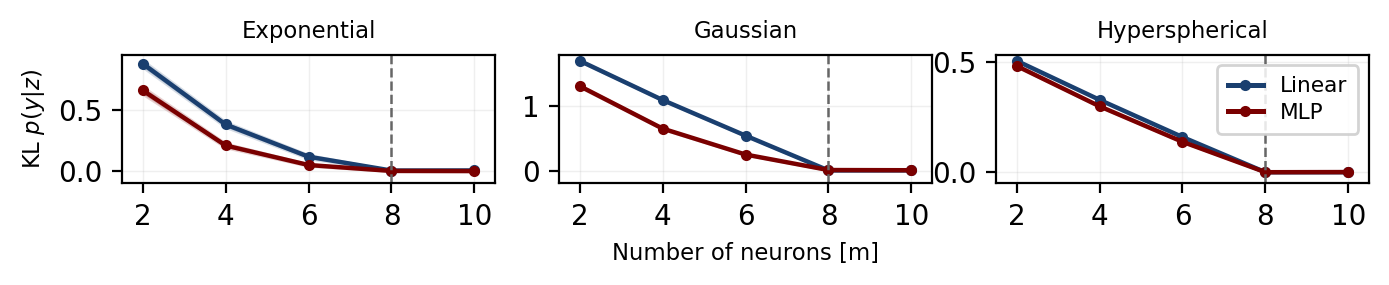}
    \caption{\textbf{Phase transition at the capacity threshold.} Final \text{KL}$(p^* \| q)$ is reported as a function of encoder dimension $m$, for linear (blue) and MLP (red) encoders across three latent distributions. The dashed vertical line marks $m = n'$, the capacity threshold.
    }
    \label{fig:nonid}
\end{figure}

We next test the \emph{undercomplete} regime ($m<n'$), in which \cref{thm:nonlinear} predicts that the global optimum of the cross-entropy loss is generically nonlinear. We fix $n'=8$ and sweep the encoder dimension $m \in \{2, 4, 6, 8, 10\}$, straddling the theoretical threshold at $m = n'$. Two encoder architectures are compared: linear encoder $h(z) = Az$ and nonlinear encoder \emph{MLP} of the same input-output dimensions. We repeat the experiment across three latent distributions. Each configuration is evaluated over ten random seeds, and we report the final $\mathrm{KL}(p^*\|q)$ as a function of $m$ in \cref{fig:nonid}.

The results reveal a clear transition at $m = n'$ that is consistent across all three distributions. When $m < n'$, the MLP achieves lower $\mathrm{KL}$ than the linear encoder across every setting, confirming that the optimal encoder is nonlinear, following \cref{thm:nonlinear}. Additionally, when $m \ge n'$, both the linear and the MLP encoder converge to near-zero $\mathrm{KL}$ divergence. This supports \cref{thm:linear_id} that sufficient capacity ensures the optimal encoder being linear to latent variables, $z$.

\subsection{Compressed Sensing Recovery}\label{sec:exp:cs}

\begin{figure}[tb]
    \centering
    \includegraphics[width=0.8\linewidth]{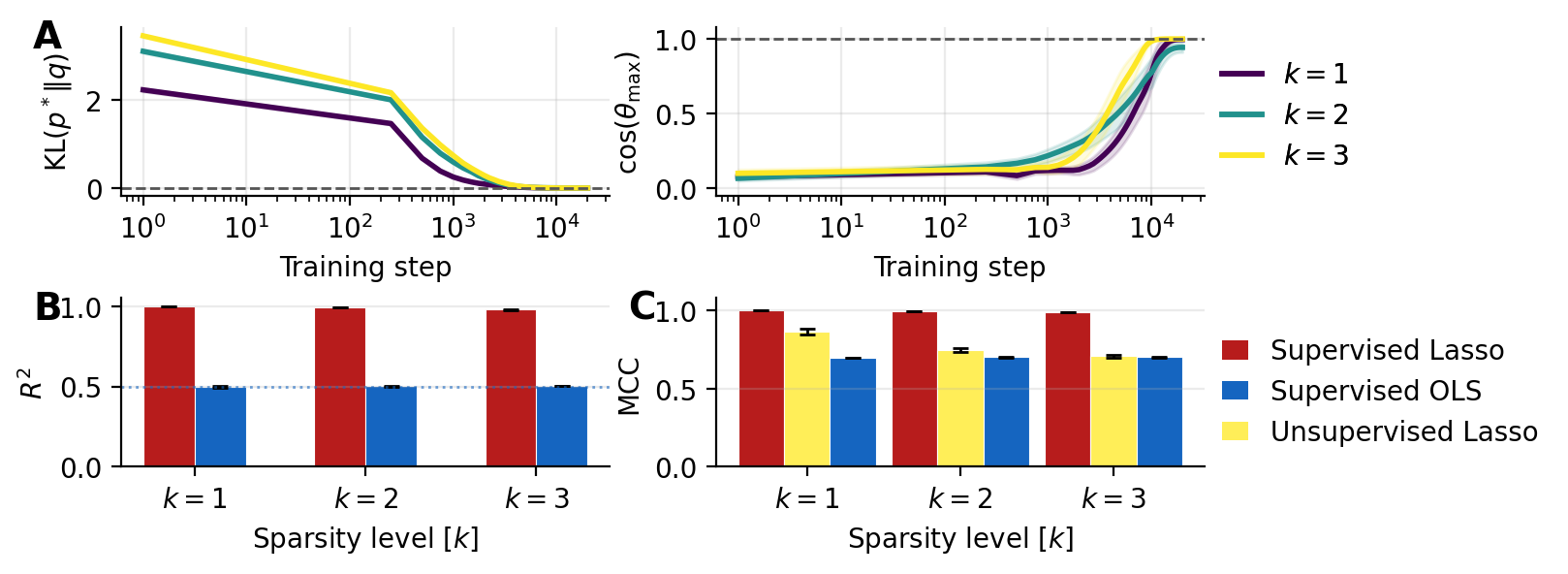}
    \caption{\textbf{Compressed Sensing Recovery.}
    \textbf{A)} Training dynamics for a linear encoder $h(z) = Az$ trained under different sparsity levels. Left: training \text{KL}$(p^* \| q)$. Right: row-space match $\cos(\theta_{\max})$ between the learned $A$ and the true $A^*$. 
    \textbf{B)} $R^2$ of supervised latent recovery by Lasso and OLS.
    \textbf{C)} MCCs of supervised latent recovery and unsupervised sparse dictionary learning. Shaded areas in \textbf{A)} and error bars in \textbf{B)} and \textbf{C)} indicate standard error of the mean over 10 random seeds.
    }
    \label{fig:cs}
\end{figure}

We then validate compressed sensing recovery in the discriminative setting, where the labels depend on sparse latents through a low-rank projection $W^* = U^*A^*$. 
The theory predicts two interlocking results: the learned encoder will precisely recover the null space of $A^*$, i.e. $\mathcal{N}(A) = \mathcal{N}(A^*)$; and this structural match enables 
full recovery of the high-dimensional sparse latent $z$ from the compressed encoding $Az$ via 
sparse decoding.

As presented in \cref{fig:cs}, we construct a discriminative DGP at full capacity $m = n'$ and sweep over three sparsity 
levels $k$ to probe different regimes while ensuring \emph{spark}-$2k$. Three recovery methods are applied to the activation
$Az$: supervised sparse inference (Lasso), dense linear decoder (OLS), and unsupervised sparse dictionary learning (Lasso). Each experiment is repeated with ten random seeds. In \cref{fig:cs}A. Training dynamics and null-space alignment demonstrate that as the $\mathrm{KL}$ 
approaches zero, the row-space alignment $\cos(\theta_{\max})$ between the learned $A$ and the 
true $A^*$ undergoes an acute transition to $1$ (\cref{fig:cs}A, right), directly confirming 
$\mathcal{N}(A) = \mathcal{N}(A^*)$ at the optimum. The recovery results complete the picture 
(\cref{fig:cs}B): supervised Lasso achieves near-perfect $R^2$ across all values of $k$, 
demonstrating faithful reconstruction of the full $n$-dimensional latent. By contrast, OLS recovery plateaus at $R^2 \approx n'/n$, confirming 
\cref{thm:cs}. Also, the MCC of the unsupervised Lasso reaches $\approx 0.86$ under extreme sparsity ($1$-sparse) and declines notably as $k$ increases, demonstrating the significant impact of sparsity on latent variable recovery and offering a principled account of why sparse autoencoders succeed in recovering latent structure from undercomplete neural representations. 

\subsection{Disentangled Neurons Generalize Across Models and Brains}\label{sec:exp:real_data}

\begin{figure}[tb]
    \centering
    \includegraphics[width=.99\linewidth]{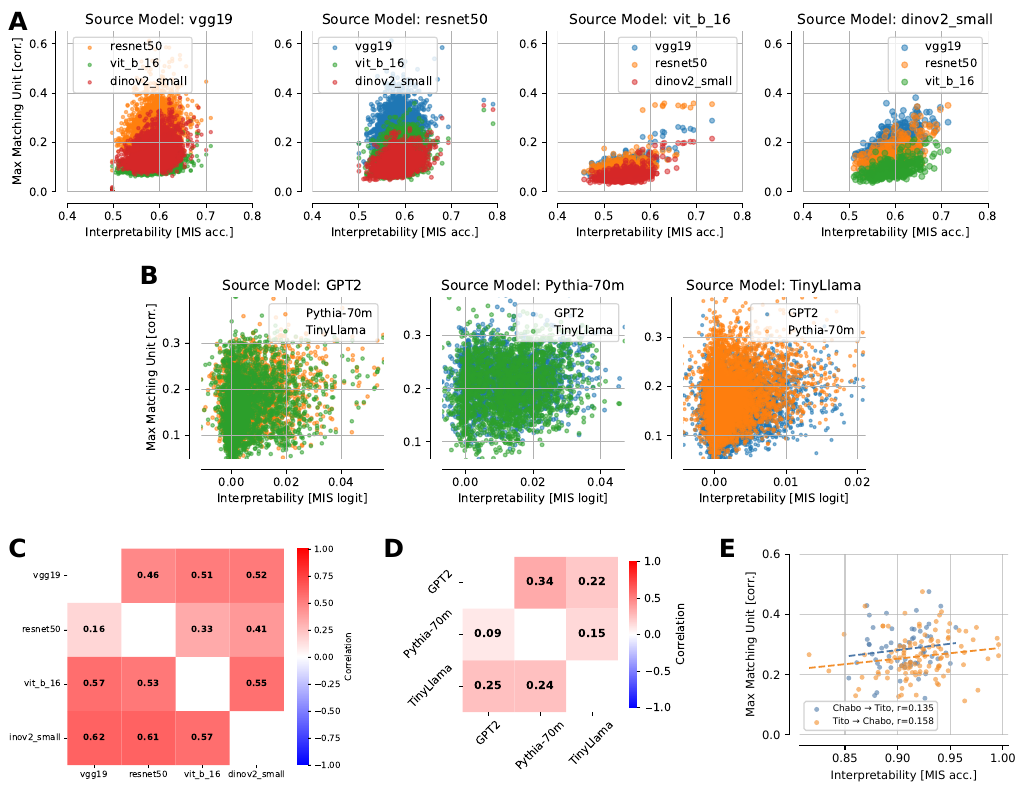}
    \caption{
        \textbf{Disentanglement predicts cross-system preservation.}
        \textbf{A,B)} Per-neuron interpretability (MIS, \citealt{zimmermann_measuring_2024}) versus the Maximum Matching Unit (MMU) score, for ImageNet-trained vision models (A) and LLMs (B; MIS logits reported as accuracy is at ceiling).
        \textbf{C,D)} MIS--MMU correlations for all model pairs in A and B.
        \textbf{E)} Directional MIS--MMU associations in macaque IT cortex \citep{majaj2015simple}. Responses to 3{,}200 images were recorded from 58 IT neurons in "Chabo" and 110 IT neurons in "Tito", two macaques.
    }
    \label{fig:real_data}
\end{figure}
\vspace{-10pt}

Our identifiability results predict that, under suitable conditions, individual neurons recover individual latent variables (\cref{thm:axis,thm:cs}). A direct consequence is that if two systems trained on similar data have each achieved this recovery, well-disentangled units in one system should have a counterpart in the other. We test this prediction across vision models, language models, and macaque cortex.
Since true latents are unobservable in real data, we use the Mechanistic Interpretability Score (MIS) of \citet{zimmermann_measuring_2024} as a proxy for disentanglement---high-MIS neurons exhibit selective, monosemantic responses consistent with having cleanly captured a latent concept. We quantify cross-system preservation via the Maximum Matching Unit (MMU) score: for each neuron, the maximum absolute correlation with any neuron in the comparison system. Our prediction is then simply a positive MIS--MMU correlation.

\paragraph{Vision models.} Across four ImageNet-trained networks (VGG-19, ResNet-50, ViT-b/16, DINOv2 ViT-S/14), MIS predicts MMU positively for all twelve directed model pairs (Fig.~\ref{fig:real_data}A,C).

\paragraph{Language models.} The same pattern, though weaker, holds across GPT-2, Pythia-70m, and TinyLlama, evaluated on the 100{,}000 most frequent Wikipedia unigrams (Fig.~\ref{fig:real_data}B,D).

\paragraph{Biological neurons.} In macaque IT cortex \citep{majaj2015simple} (macaque 1 "Chabo": 58 neurons; macaque 2 "Tito": 110 neurons; 3{,}200 natural images), the directional MIS--MMU correlations are positive. Tito $\to$ Chabo yields $r=0.158$ ($p=0.0986$;Fig.~\ref{fig:real_data}E), whereas Chabo $\to$ Tito yields $r=0.135$ with less statistical significance ($p=0.312$), plausibly due to the imbalance candidate pool of MMU.

Across all three settings, the more disentangled a unit, the more likely it is to be found in another system. This is consistent with the theoretical prediction that identifiability drives cross-system convergence on shared latent variables, although the gap between our idealized framework and real neural systems warrants caution.

\section{Limitations and Ablations}\label{sec:limitations}
Our analysis rests on a set of idealized assumptions that rarely hold exactly in real data \citep{reizinger2025empirically}. We highlight four salient ones.
\textbf{Constant conditional entropy.} The axis-alignment result (\cref{thm:axis}) relies on the class-independent entropy condition in eq.~\eqref{eq:const_entropy}. Real datasets almost always exhibit some spread in conditional entropies. In preliminary simulations, moderate deviations merely blur axis alignment rather than destroy it, suggesting robustness, but a systematic study is left for future work.
\textbf{Non-negativity.} Imposing non-negativity is analogous to restricting latents to the positive orthant of a sphere: inner products remain meaningful, but sign information is lost. Alternative sparsity models with Laplace-distributed (\emph{signed}) latents break this symmetry; \cref{thm:axis} does not apply directly there.
\textbf{Uniform class prior.} Assuming $p(y) = 1/|\mathcal{Y}|$ simplifies the inner-product identity. Unequal priors merely add a constant offset $-\log \pi_y$ to each logit row of $W^*$, and the main results extend verbatim after subtracting row means. Extreme class imbalance could nonetheless interact with optimization in ways not captured here.
\textbf{DGP.} Finally, the largest gap remains the existence of a \emph{true} sparse DGP \citep{huh2024platonic}--perhaps a philosophical question that no amount of data or theory will settle.

We also conduct ablation tests to assess the robustness of our results to violations of these assumptions. Overall, both linear identifiability and axis alignment are robust to moderate violations of constant conditional entropy, particularly with an affine readout, and to moderate departures from conditional independence and the exponential conditional distribution. In contrast, violating the non-negative latent-support assumption in the generative setting substantially degrades axis alignment. Full results and experimental details are provided in Supplement~\ref{app:axis_ablations}.

\section{Discussion}\label{sec:discussion}
Our framework unifies disparate empirical findings---dedicated, monosemantic units; polysemantic superposition; and sparse-autoencoder recovery---within a single statistical picture by combining nonlinear identifiability theory with two complementary data generating processes. Two ingredients control the qualitative regime: the \emph{causal direction} of the DGP (labels$\to$latents vs.\ latents$\to$labels) and the \emph{capacity ratio} between neurons and effective latents ($m$ vs.\ $n'$).

When capacity is sufficient ($m \geq n'$), the cross-entropy optimum identifies the latents up to an arbitrary linear transformation (\cref{thm:linear_id}). Two distinct paths then collapse this linear ambiguity to a permutation. In the generative setting, exact capacity, non-negative activations, and an energy constraint yield an axis-aligned optimum in which each neuron encodes a single latent (\cref{thm:axis}), offering one possible theoretical mechanism for grandmother-cell-like responses. In the discriminative setting, sparsity together with the spark condition on the true measurement matrix endows the encoder with a compressed-sensing geometry, so individual latents can be recovered from polysemantic representations via sparse coding (\cref{thm:cs})---an explanation for why sparse autoencoders work. When capacity is insufficient ($m < n'$), no linear encoder is optimal: the bottleneck forces nonlinear distortion and identifiability fails generically (\cref{thm:nonlinear}).

This perspective also speaks to \textit{efficient coding} \citep{barlow_single_1972,machens2005testing}, which has long struggled to reconcile distributed-coding intuitions with empirically dedicated units such as the frog's fly detector \citep{lettvin2007frog} or the mouse's bird detector \citep{zhang2012most}. The dichotomy is not between efficiency and selectivity but between regimes: under sparse, non-negative latents with sufficient capacity, selectivity \emph{is} the efficient code. Future work should relax the most restrictive assumptions---non-negativity, constant conditional entropy, the uniform class prior---and address signed or manifold-valued latents, narrowing the gap between our idealized framework and real biological and artificial systems.

\paragraph{Code availability.}
Code for reproducing the experiments and figures in this paper is available at
\url{https://github.com/klindtlab/complete-or-sparse}.

\begin{ack}

We would like to thank Luc Leh'ericy and Katrin Franke for helpful discussions.
David Klindt acknowledges the CSHL GPU cluster with assistance from the US National Institutes of Health Grant S10OD028632-01.
This work was supported by a Goodfire Research Grant and conducted using Silico, Goodfire’s interpretability research agent.

\end{ack}

\bibliographystyle{unsrtnat}
\bibliography{references}

\newpage
\section*{Supplement}

\section{Proofs}\label{app:proofs}

\subsection{Assumptions}\label{app:assum}

We delineate the assumptions for our theoretical framework in three parts: General assumptions shared across all settings, assumptions specific to the Generative setting, and assumptions specific to the Discriminative setting.

\begin{assumption}[General Assumptions]\label{ass:general}\leavevmode\par
These assumptions hold for both the Generative and Discriminative settings.
\begin{enumerate}[label=\textup{(G\arabic*)}, ref=G\arabic*]
    \item \label{ass:stimulus} \textbf{Latent-to-Data Map:}
          The data generation process is given by $x = g(z)$, where $g: \mathcal{Z} \to \mathcal{X}$ is a fixed, continuous, and invertible function. This ensures that the observation $x$ contains all information about the latent variables $z$.
    \item \label{ass:loglinear} \textbf{Log-linear True Posterior:}
          The true relationship between latents and labels is $p(y|z) = \softmax(W^* z)$ for some matrix $W^* \in \mathbb{R}^{|\mathcal{Y}| \times n}$. The effective dimensionality is $n' = \rank(W^*)$.
    \item \label{ass:learned_model} \textbf{Learned Model Structure:}
          The learner optimizes a continuous encoder $f: \mathcal{X} \to \mathcal{H}$ with hidden dimensionality \(\dim(\mathcal{H})=m\) and a linear readout $U \in \mathbb{R}^{|\mathcal{Y}| \times m}$. The composite representation is $h = f \circ g: \mathcal{Z} \to \mathcal{H}$. The model's predictive distribution is $q(y|z) = \softmax(U h(z))$.
    \item \label{ass:global_opt} \textbf{Global Optimality:}
          We assume the training procedure finds a global minimizer of the population cross-entropy loss: $\mathcal{L}(h, U) = \mathbb{E}_{Z,Y} [\mathrm{CE}(p(Y|Z), q(Y|Z))]$.
\end{enumerate}
\end{assumption}


\subsection{General Lemmas}\label{app:lemmas}

\begin{lemma}[Centering Invariance]\label{lem:centering}
Let $\bar{W} = W^* - \mathbf{1} \bar{w}^\top$ where $\bar{w} = \frac{1}{|\mathcal{Y}|} \sum_y w^*_y$ is the mean over rows. Then $\softmax(W^* z) = \softmax(\bar{W} z)$ for all $z$. Consequently, all identifiability results depend only on the centered matrix $\bar{W}$, and we assume without loss of generality that $W^*$ is row-centered.
\end{lemma}

\begin{proof}\leavevmode
For any constant vector $c \in \mathbb{R}^{|\mathcal{Y}|}$, $\softmax(v + c) = \softmax(v)$ since the shift cancels in the ratio. Setting $c = -(\bar{w}^\top z) \mathbf{1}$, we have $\softmax(W^* z) = \softmax(W^* z - (\bar{w}^\top z)\mathbf{1}) = \softmax(\bar{W} z)$.
\end{proof}

\begin{remark}
The shift-invariance of the softmax function means that the true posterior $p(y|z) = \softmax(W^* z)$ is unchanged if we add any constant vector to all rows of $W^*$. What matters for identifiability is not $W^*$ itself, but rather the space spanned by row differences $\{w^*_y - w^*_{\tilde{y}}\}_{y, \tilde{y} \in \mathcal{Y}}$. However, working with row differences directly is cumbersome. Centering $W^*$ gives us a convenient equivalence: for a row-centered matrix, $\rank(W^*) = \dim(\Span\{w^*_y - w^*_{\tilde{y}}\})$. This allows us to state identifiability results in terms of the simpler quantity $\rank(W^*)$.
\end{remark}

\begin{lemma}[Log-Ratio Identity]\label{lem:log_ratio}
Assume the true posterior is $p(y|z) = \softmax(W^* z)$ and the learned model is $q(y|z) = \softmax(U h(z))$. If $p(\cdot\mid z)=q(\cdot\mid z)$ at a given $z$, then for any pair of classes $y, \tilde{y} \in \mathcal{Y}$:
\[
    \langle u_y - u_{\tilde{y}}, h(z) \rangle = \langle w^*_y - w^*_{\tilde{y}}, z \rangle,
\]
where $u_y$ and $w^*_y$ are the $y$-th rows of $U$ and $W^*$ respectively.
\end{lemma}

\begin{proof}\leavevmode
For the given $z$, $p(y|z) = q(y|z)$ for all $y$. Writing out the softmax probabilities:
\[
    \frac{e^{\langle w^*_y, z \rangle}}{\sum_k e^{\langle w^*_k, z \rangle}} = \frac{e^{\langle u_y, h(z) \rangle}}{\sum_k e^{\langle u_k, h(z) \rangle}}.
\]
Taking the log-ratio of probabilities for classes $y$ and $\tilde{y}$ eliminates the partition functions:
\begin{align*}
    \log \frac{p(y|z)}{p(\tilde{y}|z)} &= \log \frac{q(y|z)}{q(\tilde{y}|z)} \\
    \Rightarrow\langle w^*_y - w^*_{\tilde{y}}, z \rangle &= \langle u_y - u_{\tilde{y}}, h(z) \rangle.
\end{align*}
Consequently, the identity holds throughout the support whenever the two continuous posteriors agree $p_Z$-almost surely.
\end{proof}

\begin{lemma}[Linearity from Inner Products]\label{lem:linearity}
Let $\mathcal{Z} \subseteq \mathbb{R}^n$ be a domain spanning $\mathbb{R}^n$, and let $h: \mathcal{Z} \to \mathbb{R}^m$ be a continuous function with $m \geq n'$. Suppose there exist two families of vectors $\mathcal{B} := \{b_k\}_{k \in \mathcal{K}} \subset \mathbb{R}^n$ and $\mathcal{C} := \{c_k\}_{k \in \mathcal{K}} \subset \mathbb{R}^m$ indexed by a set $\mathcal{K}$, satisfying:
\begin{equation}\label{eq:inner_prod_identity}
    \langle c_k, h(z) \rangle = \langle b_k, z \rangle, \quad \forall z \in \mathcal{Z}, \forall k \in \mathcal{K}.
\end{equation}
If $\Span(\mathcal{B})$ has dimension $n'$, then there exists a matrix $A \in \mathbb{R}^{m \times n}$ with $\rank(A) = n', \Image(A)\subseteq\Span(\mathcal C)$ such that:
\[
    h(z) = Az + \nu(z)
\]
where $\nu(z) \in \Span(\mathcal{C})^\perp$ for all $z \in \mathcal{Z}$.
\end{lemma}

\begin{proof}\leavevmode
\paragraph{Step 1: Dimension bound.}
We first show $\dim(\Span(\mathcal{C})) \geq n'$. Suppose $\sum_{k} \alpha_k c_k = 0$ for some coefficients $\{\alpha_k\}$. Taking the inner product with $h(z)$ and applying \eqref{eq:inner_prod_identity}:
\[
    0 = \Bigl\langle \sum_{k} \alpha_k c_k, h(z) \Bigr\rangle = \sum_{k} \alpha_k \langle b_k, z \rangle = \Bigl\langle \sum_{k} \alpha_k b_k, z \Bigr\rangle.
\]
Since $\mathcal{Z}$ spans $\mathbb{R}^n$, this implies $\sum_{k} \alpha_k b_k = 0$. 
Thus, every linear dependence among $\mathcal{C}$ induces a linear dependence among $\mathcal{B}$. This implies $\dim(\Span(\mathcal{C})) \geq \dim(\Span(\mathcal{B})) = n'$.

\paragraph{Step 2: Construction of $A$.}
Select indices $\mathcal{I} \subseteq \mathcal{K}$ such that $\{c_i\}_{i \in \mathcal{I}}$ forms a basis for $\Span(\mathcal{C})$. Let $C$ be the matrix with rows $c_i^\top$ for $i \in \mathcal{I}$, and let $B$ be the matrix with corresponding rows $b_i^\top$. The inner product conditions give $Ch(z) = Bz$ for all $z \in \mathcal{Z}$.

Since the rows of $C$ are linearly independent by construction, $CC^\top$ is invertible. Define:
\[
    A := C^\top (CC^\top)^{-1} B \in \mathbb{R}^{m \times n}.
\]
Let $P_C = C^\top (CC^\top)^{-1} C$ be the orthogonal projection onto $\Span(\mathcal{C})$. We decompose:
\[
    h(z) = P_C h(z) + (I - P_C) h(z) = C^\top (CC^\top)^{-1} Ch(z) + (I - P_C) h(z).
\]
Substituting $Ch(z) = Bz$:
\[
    h(z) = Az + \nu(z), \quad \text{where } \nu(z) := (I - P_C) h(z).
\]
By construction, $\Image(A)\subseteq\Span(\mathcal C)$ and $\nu(z) \in \Span(\mathcal{C})^\perp$.

\paragraph{Step 3: Rank of $A$.}
Observe that $CA = CC^\top(CC^\top)^{-1}B = B$. We now show that $\rank(B) = n'$. Since $\{c_i\}_{i \in \mathcal{I}}$ is a basis for $\Span(\mathcal{C})$, for every $c_k\in \mathcal{C}$, there exists a sequence $\{\beta_i\}_{i\in\mathcal{I}}$ such that $c_k = \sum_{i\in \mathcal{I}} \beta_ic_i$. By \cref{eq:inner_prod_identity}, we have for every $z\in\mathcal{Z}$,
\[
    \langle b_k, z\rangle = \langle c_k, h(z)\rangle
    = \Big\langle \sum_{i\in \mathcal{I}} \beta_ic_i, h(z)\Big\rangle 
    =\sum_{i\in \mathcal{I}} \beta_i\langle c_i, h(z)\rangle
    =\sum_{i\in \mathcal{I}} \beta_i\langle b_i, z\rangle
    =\Big\langle \sum_{i\in \mathcal{I}} \beta_ib_i, z\Big\rangle.
\]
Hence for all $z\in\mathcal{Z}$, $\Big\langle b_k - \sum_{i\in \mathcal{I}} \beta_ib_i, z\Big\rangle = 0$. Again, as $\mathcal{Z}$ spans $\R^n$, we have $b_k = \sum_{i\in \mathcal{I}} \beta_ib_i$. That is, $\forall k\in \mathcal{K}$, $b_k$ is a linear combination of $\{b_i\}_{i\in\mathcal{I}}$, i.e. 
\[
\Span (\mathcal{B}) = \Span (\{b_i\}_{i\in\mathcal{I}}) = \Row(B).
\]
Thus, $\rank(B) = \dim(\Span (\mathcal{B})) = n'$. For the rank of $A$: since $A$ factors through $B$, we have $\rank(A) \leq \rank(B) = n'$. Conversely, since $C$ has full row rank, $\rank(B) = \rank(CA) \leq \rank(A)$. Thus $\rank(A) = n'$.
\end{proof}

\subsection{Linear Identifiability}\label{proof:linear_id}

\begin{theorem*}[Linear identifiability]
Let \cref{ass:general} hold with $m \geq n' = \rank(W^*)$. Every global minimizer $(U, h)$ of the cross-entropy satisfies
\[
    h(z) = Az + \nu(z), \qquad UA=W^*
\]
where $A \in \mathbb{R}^{m \times n}$ has $\rank(A) = n'$, and $\nu(z) \in \mathcal{N}(U)$ for all $z$.\footnote{This does not mean that $f$ is linear, but that it inverts $g$ \textit{up to a linear transform}: $h(z) = f(g(z)) = Az + \nu(z)$.}
\end{theorem*}

\begin{remark}
The component $\nu(z)$ lies in the null space of the readout matrix $U$ and is therefore unconstrained by the classification loss; it may be nonlinear.
\end{remark}

\begin{proof}
By \cref{lem:centering}, assume without loss of generality that $W^*$ is row-centered, so $n' = \rank(W^*)$.

\paragraph{1. Establishing the inner product identity.}
The decomposition of the cross-entropy loss in equation \eqref{eq:ce_loss} gives
\[
    \mathcal L(h,U)=H(Y\mid Z)+\mathbb E_ZD_{\mathrm{KL}}\!\left(
        p(\cdot\mid Z)\,\|\,q(\cdot\mid Z)
    \right).
\]
Because $m\ge n'$, take a rank factorization
$W^*=\widetilde U\widetilde A$ with inner dimension $n'$ and pad it with
zeros to obtain
$U_0\in\mathbb R^{|\mathcal Y|\times m}$ and
$A_0\in\mathbb R^{m\times n}$ satisfying $U_0A_0=W^*$. By
\cref{ass:stimulus}, the continuous map $h_0(z)=A_0z$ is realizable by an
admissible encoder. Thus $q_0=p$ is attained. Every global minimizer consequently has zero expected KL divergence, hence
$p(\cdot\mid z)=q(\cdot\mid z)$ for $p_Z$-almost every $z$. 

We then invoke the Log-Ratio Identity (\cref{lem:log_ratio}). For any pair of classes $y, \tilde{y} \in \mathcal{Y}$:
\begin{equation}\label{eq:opt_identity}
    \langle u_y - u_{\tilde{y}}, h(z) \rangle = \langle w^*_y - w^*_{\tilde{y}}, z \rangle, \quad \forall z \in \mathcal{Z}.
\end{equation}

\paragraph{2. Linearity of the encoder.}
Define the families $\mathcal{B} := \{ w^*_y - w^*_{\tilde{y}} \}_{y, \tilde{y} \in \mathcal{Y}}$ and $\mathcal{C} := \{ u_y - u_{\tilde{y}} \}_{y, \tilde{y} \in \mathcal{Y}}$, with the natural indexing $k = (y, \tilde{y})$.

Since $W^*$ is centered with rank $n'$, the difference vectors $\mathcal{B}$ span an $n'$-dimensional subspace of $\mathbb{R}^n$. The identity \eqref{eq:opt_identity} gives $\langle c_k, h(z) \rangle = \langle b_k, z \rangle$ for all $z \in \mathcal{Z}$ and all $k$.

Applying \cref{lem:linearity} yields:
\[
    h(z) = Az + \nu(z)
\]
where $A \in \mathbb{R}^{m \times n}$ has $\rank(A) = n'$ and $\nu(z) \in \Span(\mathcal{C})^\perp$.

\paragraph{3. The residual lies in the null space of $U$.}
Since $\mathcal{C} = \{u_y - u_{\tilde{y}}\} \subseteq \mathrm{rowspace}(U)$, we have $\Span(\mathcal{C})^\perp \supseteq \mathcal{N}(U)$.

To show equality on $\nu(z)$: from $h(z) = Az + \nu(z)$ and $\nu(z) \perp \Span(\mathcal{C})$, we have for any class pair $(y, \tilde{y})$:
\[
    \langle u_y - u_{\tilde{y}}, \nu(z) \rangle = 0.
\]
This means $(U\nu(z))_y - (U\nu(z))_{\tilde{y}} = 0$ for all $y, \tilde{y}$, so $U\nu(z)$ is a constant vector. Since $U\nu(z) \in \mathrm{rowspace}(U^\top)$ and constant vectors lie in $\mathrm{rowspace}(U^\top)$ only if they are zero (by the centering of $U$ at optimality), we have $U\nu(z) = 0$, i.e., $\nu(z) \in \mathcal{N}(U)$.
\end{proof}

\subsection{Non-identifiability}\label{proof:nonlinear}

\begin{theorem*}[Non-identifiability]
Let \cref{ass:general} hold with $m < n' = \rank(W^*)$. Assume that $\supp(p_Z)$ has nonempty interior $\Omega \subseteq \R^n$, that $p_Z$ is continuous and strictly positive on $\Omega$, and that for every candidate $A \in \R^{m \times n}$ under consideration, $\mathcal{N}(A) \cap \Omega \neq \emptyset$. Then the global minimizer of the cross-entropy loss is generically nonlinear.
\end{theorem*}

\begin{remark}
The support condition is automatic in the discriminative setting where $p_Z$ has full support on $\R^n$. In the generative setting where $\supp(p_Z) = \R_+^n$, it amounts to assuming $\mathcal{N}(A)$ is not entirely confined to the boundary $\partial\R_+^n$. This holds for generic $A$ and is needed only to make the variational argument below well-posed.
\end{remark}

\begin{proof}\leavevmode
By \cref{lem:centering}, assume WLOG that $W^*$ is row-centered, so $n' = \rank(W^*)$. We proceed by contradiction: assume some linear encoder $h^*(z) = Az$ with readout $U^* \in \R^{|\mathcal{Y}| \times m}$ achieves the global minimum of $\mathcal{L}(h, U) = \E_Z[\CE(p(\cdot|Z), q_{h(Z)})]$, where $q_v := \softmax(U^* v)$. Since $m < n'$, $\dim \mathcal{N}(A) \geq n - m \geq n - n' + 1 \geq 1$.

For brevity, write $\ell(v; z) := \CE(p(\cdot|z),\, \softmax(U^* v))$, so that $\mathcal{L}(h^*, U^*) = \E_Z[\ell(h^*(Z); Z)]$. The standard softmax-plus-cross-entropy gradient is
\begin{equation}\label{eq:grad_ell}
    \nabla_v \ell(v; z) = (U^*)\T\big(\softmax(U^* v) - p(\cdot|z)\big).
\end{equation}

\paragraph{1. Pointwise stationarity on $\Omega$.}
We first show that at the global minimum,
\begin{equation}\label{eq:pw_stat}
    \nabla_v \ell|_{v = h^*(z)} = 0, \quad \forall z \in \Omega.
\end{equation}
This is the pointwise consequence of global optimality, by the fundamental lemma of the calculus of variations.

Suppose for contradiction that there exists $z_0 \in \Omega$ with $\gamma_0 := \nabla_v \ell|_{v = h^*(z_0)} \neq 0$. Since $h^*$ is continuous and $\nabla_v \ell$ is continuous in both arguments, $z \mapsto \nabla_v \ell|_{v = h^*(z)}$ is continuous. So there exists an open ball $B_\delta(z_0) \subset \Omega$ such that
\[
    \langle \nabla_v \ell|_{v = h^*(z)},\, \gamma_0\rangle \geq \tfrac{1}{2}\|\gamma_0\|^2 > 0, \quad \forall z \in B_\delta(z_0).
\]
Let $\phi: \R^n \to [0, 1]$ be a smooth bump function supported in $B_\delta(z_0)$ with $\phi(z_0) = 1$. For $\epsilon > 0$, define the continuous perturbation
\[
    \tilde h(z) := h^*(z) - \epsilon\,\phi(z)\,\gamma_0.
\]
By Taylor's theorem applied to $\ell(\cdot; z)$, which is smooth in its first argument:
\begin{align*}
    \mathcal{L}(\tilde h, U^*) - \mathcal{L}(h^*, U^*)
    &= -\epsilon \int_{B_\delta(z_0)} \phi(z) \langle \nabla_v\ell|_{h^*(z)},\, \gamma_0\rangle\, p_Z(z)\, dz + O(\epsilon^2) \\
    &\leq -\tfrac{\epsilon}{2}\|\gamma_0\|^2 \int_{B_\delta(z_0)} \phi(z)\, p_Z(z)\, dz + O(\epsilon^2).
\end{align*}
Since $p_Z > 0$ on $\Omega \supset B_\delta(z_0)$ and $\phi$ is positive on a set of positive measure, the integral is strictly positive. For $\epsilon$ small enough, $\mathcal{L}(\tilde h, U^*) < \mathcal{L}(h^*, U^*)$, contradicting global optimality. Hence \eqref{eq:pw_stat} holds.

\begin{tcolorbox}[colback=blue!10, colframe=blue!50]
\textbf{TL;DR.} Global minimization in expectation forces \emph{pointwise} stationarity in $h$ on the interior of $\supp(p_Z)$. A localized bump perturbation in any direction with nonzero gradient would strictly decrease the integral.
\end{tcolorbox}

\paragraph{2. Restriction to $\mathcal{N}(A)$.}
For $z \in \mathcal{N}(A) \cap \Omega$, $h^*(z) = Az = 0$, so $q_{h^*(z)} = \softmax(0) = \tfrac{1}{|\mathcal{Y}|}\one$. Substituting into \eqref{eq:grad_ell} and \eqref{eq:pw_stat}:
\[
    (U^*)\T\!\left(\tfrac{1}{|\mathcal{Y}|}\one - p(\cdot|z)\right) = 0, \quad \forall z \in \mathcal{N}(A) \cap \Omega.
\]
Let $u_y$ denote the rows of $U^*$ and $\bar u := \tfrac{1}{|\mathcal{Y}|} \sum_y u_y$ their mean. The condition rearranges to
\[
    \sum_y p(y|z)\, u_y = \bar u, \quad \forall z \in \mathcal{N}(A) \cap \Omega.
\]
Substituting $p(y|z) = e^{\langle w^*_y, z\rangle} / Z(z)$ and clearing the partition function $Z(z) > 0$:
\begin{equation}\label{eq:expon_id}
    \sum_y (u_y - \bar u)\, e^{\langle w^*_y, z\rangle} = 0, \quad \forall z \in \mathcal{N}(A) \cap \Omega.
\end{equation}

\begin{tcolorbox}[colback=blue!10, colframe=blue!50]
\textbf{TL;DR.} On $\mathcal{N}(A)$, the encoder collapses to zero, so the predicted distribution is uniform regardless of $U^*$. Stationarity then balances $U^*$ against the \emph{true} posterior on this subspace.
\end{tcolorbox}

\paragraph{3. Linear independence of exponentials forces $U^*$ rank-1.}
Since $\mathcal{N}(A) \cap \Omega$ is open in the subspace $\mathcal{N}(A)$ (intersection of a subspace with an open set in $\R^n$), and $\dim\mathcal{N}(A) \geq 1$, we may pick $z_* \in \mathcal{N}(A) \cap \Omega$ together with a direction $\xi \in \mathcal{N}(A)$ such that:
\begin{enumerate}[label=(\roman*)]
    \item $\xi \neq 0$ and the line segment $\{z_* + t\xi : t \in I\} \subset \mathcal{N}(A) \cap \Omega$ for some open interval $I \ni 0$ (possible since $\mathcal{N}(A)\cap\Omega$ is open in $\mathcal{N}(A)$);
    \item the scalars $\lambda_y := \langle w^*_y, \xi\rangle$ are distinct across $y$.
\end{enumerate}
Condition (ii) holds for almost every $\xi \in \mathcal{N}(A)$: it fails only on a finite union of hyperplanes within $\mathcal{N}(A)$, namely $\{\xi : \langle w^*_y - w^*_{\tilde y}, \xi\rangle = 0\}$ for $y \neq \tilde y$. (The "generic" qualifier in the theorem statement covers the degenerate cases where $W^*$ has repeated rows or where $\mathcal{N}(A)$ lies entirely in the row-difference annihilator of $W^*$. The latter cannot occur when $m < n'$ for generic $W^*$.)

Substituting $z = z_* + t\xi$ into \eqref{eq:expon_id} and defining $c_y := e^{\langle w^*_y, z_*\rangle}(u_y - \bar u) \in \R^m$:
\[
    \sum_y c_y\, e^{\lambda_y t} = 0, \quad \forall t \in I.
\]
Projecting onto each coordinate $k = 1, \ldots, m$ gives a scalar identity
\[
    \sum_y (c_y)_k\, e^{\lambda_y t} = 0, \quad \forall t \in I.
\]
Since the exponentials $\{e^{\lambda_y t}\}_y$ with distinct $\lambda_y$ are linearly independent in $C(I)$, all $(c_y)_k$ vanish. Hence $c_y = 0$ for every $y$, and since $e^{\langle w^*_y, z_*\rangle} > 0$:
\begin{equation}\label{eq:U_rank1}
    u_y = \bar u, \quad \forall y \in \mathcal{Y}.
\end{equation}

\begin{tcolorbox}[colback=blue!10, colframe=blue!50]
\textbf{TL;DR.} Sliding $z$ along a one-parameter ray in $\mathcal{N}(A) \cap \Omega$ turns the constraint into a vanishing exponential sum with distinct exponents. Linear independence forces every coefficient to vanish, collapsing $U^*$ to rank one.
\end{tcolorbox}

\paragraph{4. Contradiction via strict improvement.}
By \eqref{eq:U_rank1}, $U^* = \one\, \bar u\T$, so for every $z$:
\[
    U^* h^*(z) = \one(\bar u\T A z) = (\bar u\T A z)\one,
\]
which is a constant vector in $y$. Consequently, $q_{h^*(z)} = \softmax((\bar u\T A z)\one) = \tfrac{1}{|\mathcal{Y}|}\one$ for all $z$, and the loss is
\[
    \mathcal{L}(h^*, U^*) = \E_z\!\left[-\sum_y p(y|z)\log\tfrac{1}{|\mathcal{Y}|}\right] = \log|\mathcal{Y}|.
\]

To derive a contradiction, we exhibit a linear pair $(\tilde A, \tilde U)$ with strictly smaller loss. Define the \emph{deviation-expectation matrix} $\bar{C}\in\R^{|\mathcal{Y}|\times n}$ as follows:
\begin{align*}
\bar{C}_{y,j}&= \E_z\left[(p(y|z)-\frac{1}{|\mathcal{Y}|})z_j\right] \\
&= \E_z\left[p(y|z)z_j\right] - \frac{1}{|\mathcal{Y}|}\E_zz_j
\end{align*}
As $\sum_y \bar{C}_{y,j} = \E_z\left[\sum_yp(y|z)z_j\right] - \sum_y\frac{1}{|\mathcal{Y}|}\E_zz_j = \E_zz_j - \E_zz_j = 0$, we have $\bar{C}$ is row centralized. Let $\bar{C} = P\Sigma Q^\top$ be the SVD decomposition of $\bar{C}$. By Eckart–Young–Mirsky theorem, the best $m$ rank approximation of $\bar{C}$ is $\bar{C}_m = P_m\Sigma_mQ^\top_m$. Then we define $(A_\epsilon, U_\epsilon)$ as:
\begin{align*}
    U_\epsilon &= \frac{\sqrt{\epsilon}}{\|\bar{C}_m\|_F^{1/2}} P_m\Sigma_m^{1/2} \\
    A_\epsilon &= \frac{\sqrt{\epsilon}}{\|\bar{C}_m\|_F^{1/2}} \Sigma_m^{1/2} Q^\top_m.
\end{align*}
Denote $V = \bar{C}_m / \|\bar{C}_m\|_F $, thus $\epsilon V = U_\epsilon A_\epsilon$. As $\bm{1}^\top \bar{C} =0 \Leftrightarrow \bm{1}^\top P_{[:,j]}=0$, it directly gives $\bm{1}^\top V = 0$. Now we write out the loss function for $(A_\epsilon, U_\epsilon)$.
\begin{align*}
f(\epsilon) &= \mathcal{L}(A_\epsilon, U_\epsilon) \\
&= \E_z\left[-\sum_yp(y|z)\log \softmax(\epsilon V z)_y\right] \\
&= \E_z\left[-\epsilon\sum_yp(y|z)( V z)_y + \log\sum_{y}e^{\epsilon(Vz)_{y}}\right]
\end{align*}
Here, notice that when $\epsilon =0$, we have $q(y|z) = 1/|\mathcal{Y}|$ uniform and $\mathcal{L}(U_0,A_0) = \log |\mathcal{Y}|$. Therefore, to prove contradiction, we need only prove $f'(0)<0$ equivalently.
\begin{align*}
    f'(0) &= \left.\frac{\partial}{\partial\epsilon}f\right|_{\epsilon = 0} \\
    &= \left.\E_z\left[-\sum_yp(y|z)(Vz)_y + \frac{\sum_y(Vz)_ye^{\epsilon(Vz)_y}}{\sum_ye^{\epsilon(Vz)_y}}\right]\right|_{\epsilon = 0} \\
    &= \E_z\left[-\sum_yp(y|z)(Vz)_y + \frac{1}{|\mathcal{Y}|}\bm{1}^\top Vz\right] \\
    &= \E_z\left[-\sum_yp(y|z)(Vz)_y\right] \\
    &= -\sum_{y,j}V_{y,j}\E_z\left[p(y|z)z_j\right] \\
    &= -\sum_{y,j}V_{y,j}\left(\E_z\left[p(y|z)z_j\right] - \frac{1}{|\mathcal{Y}|}\E_zz_j\right) -\sum_{y,j}V_{y,j}\cdot \frac{1}{|\mathcal{Y}|}\E_zz_j \\
    &= -\langle V, \bar{C}\rangle_F - \frac{1}{|\mathcal{Y}|}\sum_{j} \E_zz_j\sum_yV_{y,j} \\
    &= -\|\bar{C}_m\|_F < 0
\end{align*}

\begin{tcolorbox}[colback=blue!10, colframe=blue!50]
\textbf{TL;DR.} Once $U^*$ collapses to rank one, the network outputs uniform predictions and incurs the entropy bound $\log|\mathcal{Y}|$. Any linear encoder that “shifts” the latent variable $z$ in the direction of the difference between the true distribution and the uniform distribution, i.e. $p(y|z) - 1/|\mathcal{Y}|$, will achieve strictly lower loss.
\end{tcolorbox}

This completes the contradiction, so no linear encoder achieves the global minimum.

\end{proof}

\subsection{Permutation Identifiability}

\subsubsection{Axis-Aligned Neurons (Generative Setting)}\label{proof:axis}

\paragraph{Additional Assumptions}

\begin{assumption}[Generative Setting Assumptions]\label{ass:gen}\leavevmode\par
These assumptions apply to the generative setting ($Y \to Z$) where we seek axis-alignment.
\begin{enumerate}[label=\textup{(GEN\arabic*)}, ref=GEN\arabic*]
    \item \label{ass:nonneg} \textbf{Non-negative Assumptions:}
          The latent space is $\mathcal{Z} = \mathbb{R}_+^n$. The activation space is $\mathcal{H} = \mathbb{R}_+^m$, e.g., from ReLU networks.
    \item \label{ass:exponential} \textbf{Exponential Conditionals:}
          The true conditional $p(z|y)$ follows a factorized exponential family:
          \begin{equation*}
              p(z|y) = \prod_{i=1}^{n} \lambda_i(y) \exp\bigl(-\lambda_i(y) z_i\bigr).
          \end{equation*}
    \item \label{ass:entropy} \textbf{Constant Conditional Entropy:}
          The product $\prod_{i=1}^n \lambda_i(y)$ is independent of $y$. Combined with a uniform prior $p(y) = 1/|\mathcal{Y}|$, this yields $p(y|z) = \softmax(W^* z)$ with $n' = n$.
    \item \label{ass:const_var}
    \textbf{Fixed within-class population variance:}
          For a constant $C>0$ shared by all admissible representations,
          \begin{equation}\label{eq:within_variance}
              \mathcal V_{\mathrm{within}}(h)
              :=\E_Y\!\left[
                  \Tr\operatorname{Cov}(h(Z)\mid Y)
              \right]
              =\E\!\left[
                  \|h(Z)-\E[h(Z)\mid Y]\|_2^2
              \right]
              =C.
          \end{equation}
          
    \item \label{ass:energy} \textbf{Energy Constraint:}
          Among global minimizers satisfying \eqref{eq:within_variance}, the learner selects the solution minimizing $\mathcal{R}(h) = \E_z\|h(Z)\|_2^2$.
\end{enumerate}
\end{assumption}

\begin{theorem*}[Axis-aligned neurons]
Following \cref{ass:general} and \cref{ass:gen}, in the generative setting with $m=n=n'=\operatorname{rank}(W^*)$, let $(U, h)$ be any minimizer of $\mathcal R(h) = \E_z\|h(Z)\|_2^2$ among the cross-entropy global minimizers. Then
$$
h(z)=DP_\sigma z,
$$
where $P_\sigma$ is a permutation matrix and $D=\operatorname{diag}(\beta_1,\ldots,\beta_n)$ with $\beta_j>0$. That is, the representation is identifiable up to permutation and component-wise positive scaling.
\end{theorem*}

\begin{proof}\leavevmode

We divide the proof into four steps.
\paragraph{1. Pure linear representation under exact capacity.}
By Theorem \ref{thm:linear_id}, every global minimizer of the cross-entropy satisfies
\[
    h(z)=Az+\nu(z),
    \qquad
    \nu(z)\in\Ker(U),
    \qquad
    UA=W^*,
\]
where $A\in\mathbb R^{n\times n}$ has rank $n$. Since $m=n'=n$,
\[
    n=\rank(W^*)=\rank(UA)
      \leq\rank(U)\leq m=n.
\]
Thus $U$ has full column rank and $\Ker(U)=\{0\}$. Hence $\nu(z)=0$ for every
$z$, so $h(z)=Az$, with $A$ invertible.

\paragraph{2. The linear map is entrywise non-negative.}
By \ref{ass:nonneg}, $Az=h(z)\in\mathbb R_+^n$ for every
$z\in\mathbb R_+^n$. In particular, by evaluating on the coordinate rays, each column $Ae_i$ is non-negative. Therefore
\[
    A_{ji}\geq0
    \qquad\text{for every }j,i\in[n].
\]

\paragraph{3. Attainable lower bound of $\mathcal{R}(h)$.}

Let $r_i(y):=\lambda_i(y)^{-1}>0, r(y):=(r_1(y),\ldots,r_n(y))^\top.$ Under \ref{ass:exponential},
\[
    \E[Z\mid Y=y]=r(y),
    \qquad
    \operatorname{Cov}(Z\mid Y=y)
    =\diag \!\left(r_1(y)^2,\ldots,r_n(y)^2\right).
\]
Consequently,
\begin{align}
    \mathcal V_{\mathrm{within}}(h)
    &=\E_Y\Tr\!\left(
        A\operatorname{Cov}(Z\mid Y)A^\top
      \right)\notag\\
    &=\sum_{j=1}^n\sum_{i=1}^nA_{ji}^2c_i=C,
    \qquad
    c_i:=\E_Y[r_i(Y)^2]>0.
    \label{eq:gen4_diagonal}
\end{align}

The conditional bias--variance identity of \ref{ass:energy} and
\eqref{eq:gen4_diagonal} give
\begin{align}
    \mathcal R(A)
    &= \E_z\|AZ\|^2_2\notag\\
    &=\E_Y\!\left[
        \Tr\operatorname{Cov}(AZ\mid Y)
        +\|\E[AZ\mid Y]\|_2^2
      \right]\notag\\
    &=C+\E_Y\|A\E[Z\mid Y]\|_2^2\notag\\
    &=C+\sum_{j=1}^n\sum_{i,k}
        A_{ji}A_{jk}\E_Y[r_i(Y)r_k(Y)]\notag\\
    &=2C
      +2\sum_{j=1}^n\sum_{1\leq i<k\leq n}
        A_{ji}A_{jk}\E_Y[r_i(Y)r_k(Y)].
    \label{eq:axis_energy}
\end{align}
Because for all $j,i\in[n]$, $A_{ji}\geq0$ and $r_i(y) > 0$, every term added to $2C$ is non-negative. Hence $\mathcal R(A)\geq2C$.

It remains essential to show that this lower bound is feasible under all
constraints. Let
\[
    \alpha_i:=\sqrt{\frac{C}{nc_i}},
    \qquad
    A_0:=\diag(\alpha_1,\ldots,\alpha_n),
    \qquad
    U_0:=W^*A_0^{-1}.
\]
Then $A_0$ is non-negative and invertible,
\[
    \sum_{i=1}^n\alpha_i^2c_i=C,
    \qquad
    U_0A_0=W^*,
\]
so $(h_0(z)=A_0z,U_0)$ is an admissible global minimizer and
$\mathcal R(A_0)=2C$. Hence the minimum activation energy is attained and is equal to $2C$.

\paragraph{4. Equality forces a positive scaled permutation.}
Let $A_\star$ denote the linear map of any activation-energy minimizer. Equality
in \eqref{eq:axis_energy} and strict positivity of every $r_i(y)$ imply
\[
    (A_\star)_{ji}(A_\star)_{jk}=0
    \qquad
    \text{for every }j\in[n]\text{ and }i\ne k.
\]
Thus every row of $A_\star$ is at most one-sparse. Since $A_\star$ is an invertible
$n\times n$ matrix, every row is nonzero and two rows cannot use the same
column. Therefore every row and every column contains exactly one positive
entry. For a permutation $\sigma$ and positive numbers
$\beta_j:=(A_\star)_{j,\sigma(j)}$,
\[
    A_\star=DP_\sigma,
    \qquad
    D=\diag(\beta_1,\ldots,\beta_n),
\]
and hence $h(z)=DP_\sigma z$.
\end{proof}

\subsubsection{Compressed Sensing Recovery (Discriminative Setting)}\label{proof:cs}

\paragraph{Additional Assumptions}

\begin{assumption}[Discriminative Setting Assumptions]\label{ass:dis}\leavevmode\par
These assumptions apply to the discriminative setting ($Z \to Y$) where we analyze compressed sensing recovery.
\begin{enumerate}[label=\textup{(DIS\arabic*)}, ref=DIS\arabic*]
    \item \label{ass:factorization} \textbf{Low-Rank Factorization:}
          The true weight matrix factorizes as $W^* = U^* A^*$, where $A^* \in \mathbb{R}^{n' \times n}$ is a measurement matrix and $U^* \in \mathbb{R}^{|\mathcal{Y}| \times n'}$ is a readout matrix. Since $n'=\rank(W^*)$, this implies the task depends only on $A^* z$.
    \item \label{ass:sparsity} \textbf{Latent Sparsity:}
          The latent variables are $k$-sparse: $\|z\|_0 \leq k$ for all $z \in \mathrm{supp}(p_Z)$.
    \item \label{ass:spark} \textbf{Sparse injectivity:}
          The true operator satisfies $\operatorname{spark}(A^*)>2k$, equivalently,
          \[
              \Ker(A^*)\cap
              \{v\in\mathbb R^n:\|v\|_0\leq2k\}
              =\{0\}.
          \]
\end{enumerate}
\end{assumption}

\begin{theorem*}[Compressed sensing recovery]
Following \cref{ass:general} and \cref{ass:dis}, in the discriminative setting with $W^* = U^* A^*$, assume $m\geq n'$, latents are $k$-sparse, and $\operatorname{spark}(A^*)>2k$. Let $P_U$ denote the
orthogonal projector onto $\Row(U)$ and define the classifier-visible
representation $\widetilde h(z):=P_Uh(z)=Az$. Then the learned encoder satisfies $\mathcal{N}(A) = \mathcal{N}(A^*)$. Consequently,
$\operatorname{spark}(A)=\operatorname{spark}(A^*)>2k$, and $z$ can be recovered from the unique $k$-sparse solution of
\[
    Az=\widetilde h(z).
\]
\end{theorem*}

\begin{proof}\leavevmode
By \cref{lem:centering}, assume without loss of generality that $W^*$ is row-centered.

\paragraph{1. Linear identifiability.}
Since $W^* = U^* A^*$ with $\rank(A^*) = n'$, we have $\rank(W^*) = n'$. With $m \geq n'$, \cref{thm:linear_id} implies the learned encoder is linear:
\[
    h(z) = Az + \nu(z)
\]
with $\rank(A) = n'$ and $\nu(z) \in \mathcal{N}(U)$.

\paragraph{2. Matching nullspaces.}
We show $\mathcal{N}(A) = \mathcal{N}(A^*)$. By the Log-Ratio Identity (\cref{lem:log_ratio}), for any $y, \tilde{y} \in \mathcal{Y}$:
\[
    \langle u_y - u_{\tilde{y}}, h(z) \rangle = \langle w^*_y - w^*_{\tilde{y}}, z \rangle.
\]
Since $h(z) = Az$ and $W^* = U^* A^*$, we can rewrite both sides:
\begin{align*}
    \text{LHS} &= \langle u_y - u_{\tilde{y}}, Az+\nu(z) \rangle = (u_y - u_{\tilde{y}})^\top A z + (u_y - u_{\tilde{y}})^\top\nu(z) = (u_y - u_{\tilde{y}})^\top A z \\
    \text{RHS} &= \langle w^*_y - w^*_{\tilde{y}}, z \rangle = (w^*_y - w^*_{\tilde{y}})^\top z = (u^*_y - u^*_{\tilde{y}})^\top A^* z,
\end{align*}
where $\nu(z) \in \mathcal{N}(U)$. Now suppose $z \in \mathcal{N}(A)$, i.e., $A z = 0$. Then the LHS vanishes for all class pairs $(y, \tilde{y})$. Hence, we have $U^*A^* z = c\bm{1}$. Additionally, since $\bm{1}^\top U^*A^* = \bm{1}^\top W^* = 0$ and $A^*$ has full row rank, we have $\bm{1}^\top U^*=0$. Thus,
\[
0 = \bm{1}^\top U^*A^*z = c|\mathcal{Y}|
\]
That is, $c = 0$ and $A^*z \in \mathcal{N}(U^*)$. As $n'= \rank (W^*) = \rank (U^*A^*)\le \rank(U^*)\le n'$, this means that $U^*$ is full column rank and $\mathcal{N}(U^*) = \{0\}$. Thus $A^*z = 0$ and
\[
    \forall z\in\mathcal{N}(A), z\in \mathcal{N}(A^*) \Longrightarrow \mathcal{N}(A) \subseteq \mathcal{N}(A^*).
\]
Conversely, both $A \in \mathbb{R}^{m \times n}$ and $A^* \in \mathbb{R}^{n' \times n}$ have rank $n'$, so:
\[
    \dim(\mathcal{N}(A)) = n - n' = \dim(\mathcal{N}(A^*)).
\]
Combined with the inclusion above, we conclude $\mathcal{N}(A) = \mathcal{N}(A^*)$.

\paragraph{3. Factorization $A = BA^*$.}
Since $A$ and $A^*$ have the same nullspace of rank $n'$, their row spaces are identical. This means there exists an left invertible matrix $B \in \mathbb{R}^{m \times n'}$ such that:
\[
    A = B A^*.
\]
To see this explicitly: let $A^* = Q R$ be a QR decomposition where $Q \in \mathbb{R}^{n' \times n'}$ is orthogonal and $R \in \mathbb{R}^{n' \times n}$ has the same row space as $A^*$. Since $A$ has the same row space, we can write $A = \tilde{B} R$ for some full-rank $\tilde{B} \in \mathbb{R}^{m \times n'}$. Then $A = \tilde{B} Q^\top Q R = (\tilde{B} Q^\top) A^*$, giving $B = \tilde{B} Q^\top$.

\paragraph{4. Sparse recovery.}
For any two distinct $k$-sparse vectors $z_1, z_2$ satisfying $h(z_1) = h(z_2)$, consider their difference $\Delta = z_1 - z_2$, which is at most $2k$-sparse. We have:
\[
    h(z_1) = h(z_2) \iff A z_1 = A z_2 \iff A \Delta = 0 \iff B A^* \Delta = 0.
\]
Since $B$ is invertible, $B A^* \Delta = 0$ if and only if $A^* \Delta = 0$. As $\operatorname{spark}(A^*)>2k$, no nonzero $2k$-sparse vector lies in $\mathcal{N}(A^*)$. Therefore $\Delta = 0$, i.e., $z_1 = z_2$.

Thus $h$ is injective on $k$-sparse vectors. Given $(U, h(z))$, the latent $z$ can be recovered via standard sparse coding algorithms (e.g., Basis Pursuit, Lasso, or OMP), which find the sparsest solution to $Az = \widetilde h(z)$.
\end{proof}

\begin{corollary*}[Basis Pursuit recovery]\label{cor:basis_pursuit}
Following \cref{thm:cs}, suppose additionally that
$A^*$ satisfies the Null Space Property (NSP) of order $k$. Then $A$
satisfies the same property as $\mathcal{N}(A)=\mathcal{N}(A^*)$. Consequently, every
$k$-sparse latent $z$ is the unique solution of
\[
    \min_{z\in\mathbb R^n}\|z\|_1
    \quad\text{subject to}\quad
    Az=\widetilde h(z).
\]
\end{corollary*}

\begin{proof}
The Null Space Property depends only on the null space. It is
therefore inherited by $A$ from $A^*$, and the conclusion follows from the
standard equivalence between NSP of order $k$ and uniform exact Basis-Pursuit
recovery of all $k$-sparse vectors.
\end{proof}

\section{Experimental Details}\label{app:exp_details}

\paragraph{Hardware \& Software.}  
All simulations were executed on a single \textit{NVIDIA Tesla V100} (32 GB) GPU.  
The software environment comprised Python \texttt{3.12.3}, PyTorch \texttt{2.2.2} compiled against CUDA \texttt{12.1} and cuDNN \texttt{8902}, and scikit-learn \texttt{1.6.1}.  

\paragraph{Random Seeds.}  
Every figure and table reports the mean $\pm$ s.e.m.\ over ten independent runs.  
We deterministically set \texttt{seed}$=42+i$ with $i\in\{0,\dots,9\}$, forwarding the value to NumPy, PyTorch (CPU and CUDA), and Python’s \texttt{random} module.

\paragraph{Hyper-parameters.}  
All optimiser and model settings were adopted verbatim from the study by \citet{reizinger2024cross} (see their Sec.~4.2 and App.~B).  
Re-using this validated grid eliminates the risk of inadvertently tuning hyper-parameters to our synthetic benchmarks.

\subsection{Axis-aligned Neurons ($m=n'$)}\label{app:exp:axis}

\begin{figure}[tb]
    \centering
    \includegraphics[width=0.9\linewidth]{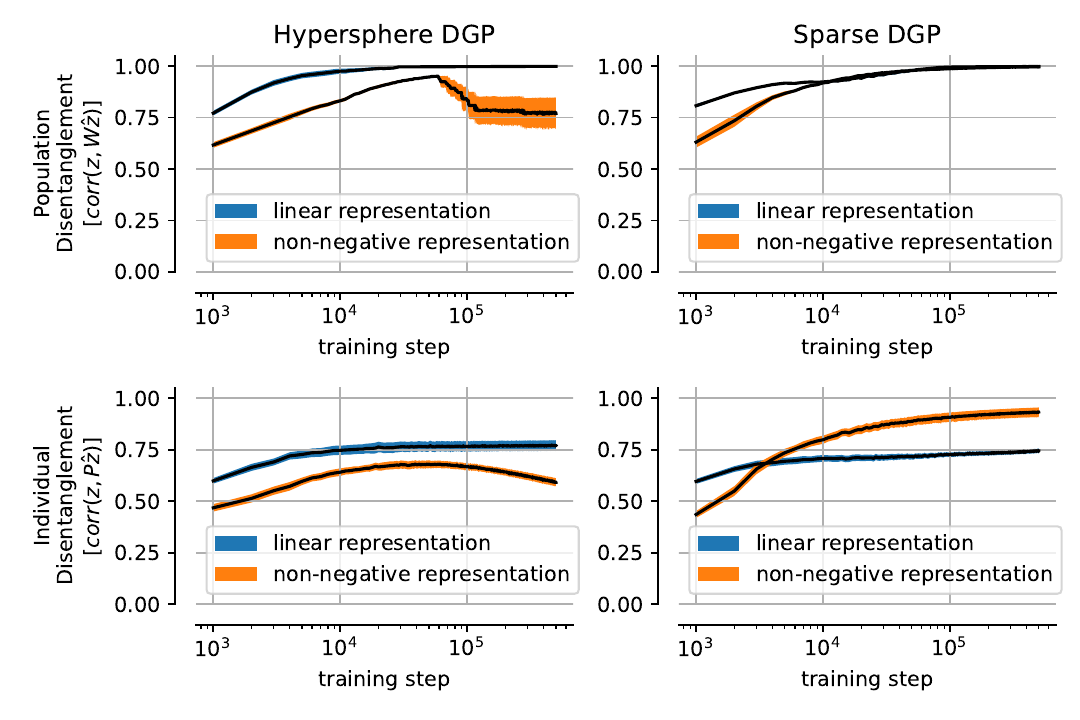}
    \caption{
    \textbf{Empirical Results.}
    \textit{Same as in main text but with exponential instead of ReLU nonlinearity, which is easier to optimize in this setting.}
    \textbf{Top left)} DGP with latents on a hypersphere \citep{zimmermann_contrastive_2021,reizinger2025empirically}, training a neural network with a linear representation (i.e., no final activation function, blue) and a non-negative representation (i.e., scale normalization and exponential nonlinearity as final activation function, orange); Disentanglement is measured up to arbitrary linear transformations.
    \textbf{Bottom left)} same as top left, but disentanglement is measured only up to permutations, measuring axis-aligned latent variable recovery.
    \textbf{Top right)} Our proposed sparse DGP with exponential distributions over non-negative latent variables, same architectures as on the left (with/without activation function), linear disentanglement reported.
    \textbf{Bottom right)} same as top right but individual, axis-aligned disentanglement measured. Shaded areas in all plots indicate standard error of the mean over 10 random seeds.
    }
    \label{fig:lin_rep}
\end{figure}

\begin{figure}[t]
    \centering
    \includegraphics[width=1\linewidth]{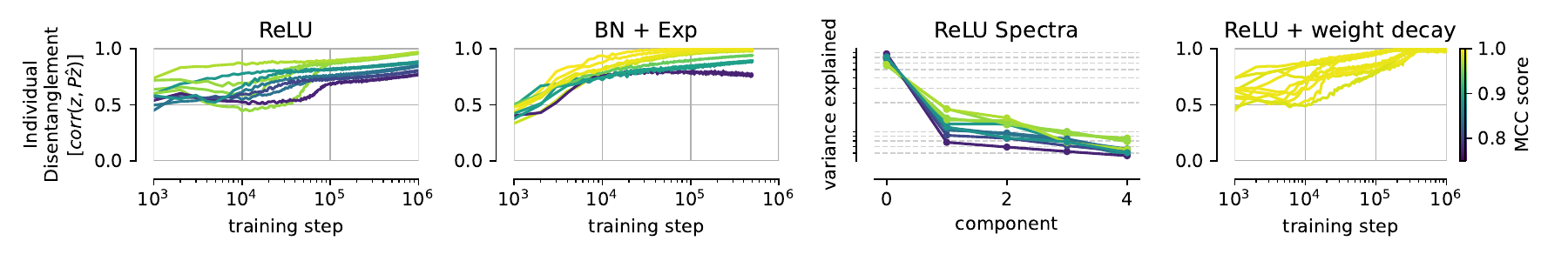}
    \caption{\textbf{Optimization Dynamics.}
    \textbf{A)} MCC curves over 10 random seeds.
    \textbf{B)} Same as A, but with batch normalisation and exponential nonlinearity.
    \textbf{C)} ReLU models with better final performance have a flatter power spectrum.
    \textbf{D)} Adding weight decay \(0.03\) improves convergence for ReLU model.
    }
    \label{fig:optimization}
\end{figure}

For the \textit{full-capacity} experiments we considered five latent variables.  
In the sparse condition each component followed an exponential distribution with rate~$\lambda=10$, whereas in the hyperspherical condition latents were drawn from a von Mises–Fisher distribution with concentration $\kappa=150$ and subsequently normalised to unit length.  
Latents were assigned to one of $C=100$ class prototypes, obtained once per seed by sampling $C$ random unit vectors and performing nearest-neighbour assignment.  
Every configuration used $N=10\,000$ latent–label pairs.

A fixed, invertible forward map $g$—implemented as an MLP with layer widths $5\!\rightarrow\!64\!\rightarrow\!64\!\rightarrow\!64\!\rightarrow\!5$, \textit{SiLU} activations, and orthonormal weight initialisation—generated the observations.  
The encoder $f$ mirrored this architecture.  Its output layer was either left \textit{linear} or constrained to be \textit{non-negative}.  In the main text we achieved non-negativity with a ReLU; in Fig.~\ref{fig:lin_rep} of the supplement we instead applied an element-wise exponential followed by scale-only batch normalisation (denoted \textit{exp+BN}), which improves gradient flow in this toy setting.  
A linear classifier $W\!: \mathbb{R}^{5}\!\rightarrow\!\mathbb{R}^{100}$ was trained jointly with $f$ under a cross-entropy objective.

Optimisation used Adam with default $\beta$-parameters $(0.9,0.999)$, $\epsilon=10^{-8}$, a learning-rate of $3\times10^{-4}$, batch size 512, and no weight decay unless explicitly noted.  Training proceeded for $50\,000$ gradient steps (approximately 500 epochs).

We quantified \textit{linear alignment} by the $R^{2}$ of an ordinary-least-squares fit $z\!\leftarrow\!h(z)$, averaged across dimensions.  
\textit{Axis alignment} was measured by the mean-correlation coefficient (MCC) of \citet{hyvarinen2016tcl}, obtained after Hungarian matching between recovered axes and ground-truth latents.

\paragraph{Optimization Dynamics.}\label{app:opt_dynamics}
As shown in \cref{fig:lin_rep_relu}, although the non-negative encoder eventually resolves the axes, its convergence is noticeably slower. 
We wanted to know what causes the very slow convergence, beyond the fact that it has to converge to a permutation.
In Fig.~\ref{fig:optimization}B we tried using a different activation function (exponential) and scale (batch) normalization to alleviate potential gradient issues in the ReLU activation.
While this seemed to improve convergence in many cases, it is not ideal because, in contrast to ReLU, this is not a common activation function so that insights in this setting are less likely to bear on real systems.
In a different line of attack, we observed in Fig.~\ref{fig:optimization}C that the models which converged to a perfect solution tended to have a flatter power spectrum.
This motivated the hypothesis that some penalty on the activation scale might be important--which is also an important assumption in prior work to obtain axis-alignment \citep{whittington2022disentanglement}.
Again, in the spirit of using relevant building blocks, we tried a weight decay of \(0.03\) which did indeed lead to faster and more reliable convergence across all seeds (Fig.~\ref{fig:optimization}D).

To probe the optimisation dynamics, we varied only three factors relative to the base setting: the output activation (ReLU versus exp+BN), the $L_{2}$ weight-decay coefficient $\lambda_{\mathrm{WD}}\in\{0,0.01,0.03\}$, and—when studying gradient-flow effects—replacement of ReLU by exp+BN.  
Power-spectrum slopes (panel C) were computed from the singular values of the centred activation matrix $\hat{Z}=h(G(Z))-\mathbb{E}[\,h(G(Z))\,]$ evaluated on the validation set. 


\begin{figure}[t]
    \centering
    \includegraphics[width=\linewidth]{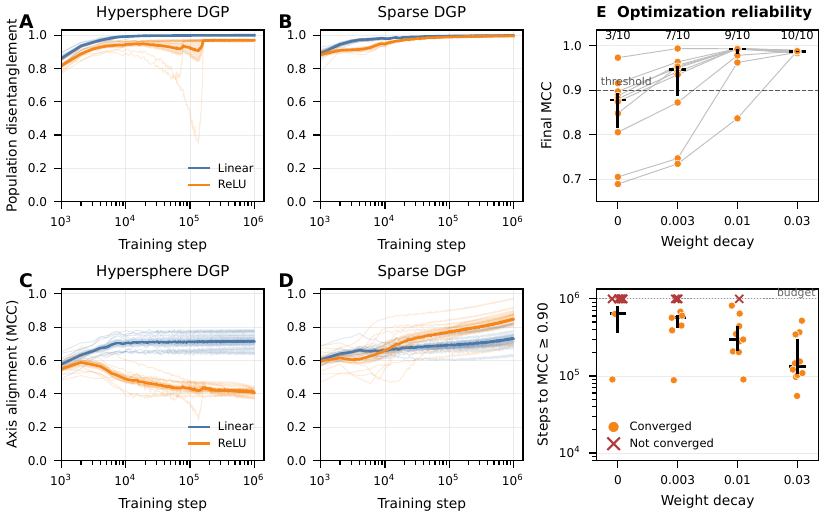}
    \caption{
    \textbf{Seed-level optimization dynamics and finite-budget reliability.}
    \textbf{A--D)} Exact-capacity experiments under a hyperspherical DGP
    (A,C) and the class-conditional sparse exponential DGP (B,D), using a
    linear output (blue) or a non-negative ReLU output (orange). A,B show
    alignment up to an unrestricted linear map, whereas C,D show
    permutation-and-scaling axis alignment measured by MCC. Thin lines denote
    individual initialization seeds, thick lines denote their mean, and shaded
    bands denote \(\pm\) s.e.m.\ over ten seeds. All A--D runs use zero weight
    decay.
    \textbf{E)} Paired weight-decay sweep for the sparse-ReLU condition. The
    upper panel shows final MCC; orange points denote individual runs, gray
    lines connect the same initialization across weight-decay values, and the
    counts report the number of runs reaching MCC \(\geq 0.90\). The lower
    panel shows the first update at which this threshold was reached. Black
    bars denote the median and interquartile range. Red crosses denote runs that
    did not reach the threshold within the \(10^6\)-update budget.
    }
    \label{fig:optimization_reliability}
\end{figure}

Individual trajectories in
Fig.~\ref{fig:optimization_reliability}A--D show that the sparse-ReLU encoder
maintains high linear alignment but can approach axis alignment slowly and
with substantial initialization dependence. Without weight decay, only
\(3/10\) runs reached MCC \(\geq0.90\) within \(10^6\) updates, and final MCC
had median \(0.878\) [IQR \(0.816,0.895\)]. At weight decay \(0.003\),
\(0.01\), and \(0.03\), the corresponding success counts were \(7/10\),
\(9/10\), and \(10/10\), and median final MCC was \(0.946\), \(0.993\), and
\(0.987\), respectively. Among runs that reached the threshold, median
time-to-threshold decreased from \(640{,}000\) updates without weight decay to
\(569{,}000\), \(297{,}000\), and \(133{,}500\) updates at weight decay
\(0.003\), \(0.01\), and \(0.03\), respectively.  Thus, weight decay substantially improves the finite-budget reliability and
speed of reaching highly axis-aligned solutions in this experiment, which suggests an optimization option for practice.

\paragraph{Assumption Ablations.}\label{app:axis_ablations}

We next perturb the assumptions underlying \cref{thm:axis} one family at a
time. Based on the previous results, we use weight decay \(0.03\) as the stable reference
configuration for these ablations except for the energy-regularization panel. Each condition is trained for
\(10^6\) updates using ten paired initialization seeds same as the optimization dynamics analysis. Within each comparison,
the class-scale vectors, nonlinear mixing function, training-batch seed, and
evaluation set are held fixed.

\begin{figure}[t]
    \centering
    \includegraphics[width=\linewidth]
    {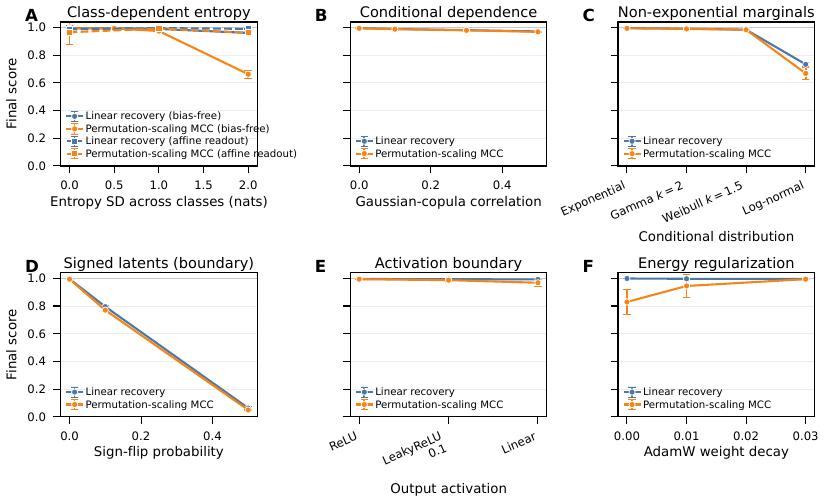}
    \caption{
    \textbf{Sensitivity to the assumptions of the axis-alignment result.}
    Each point shows mean \(\pm\) s.e.m.\ over ten paired initialization seeds
    after \(10^6\) updates. Blue denotes held-out alignment up to an
    unrestricted linear map, and orange denotes permutation-and-scaling MCC.
    \textbf{A)} Class-dependent conditional entropy, with both bias-free and
    affine-readout controls.
    \textbf{B)} Conditional dependence introduced through a Gaussian copula
    while preserving the exponential marginals.
    \textbf{C)} Alternative class-conditional marginal families matched in
    conditional mean.
    \textbf{D)} Signed latent support introduced by independently flipping
    latent signs.
    \textbf{E)} Relaxations of output non-negativity using LeakyReLU and a
    linear output.
    \textbf{F)} AdamW weight decay. All other data-generation, training, and
    evaluation randomness is shared across paired conditions.
    }
    \label{fig:axis_assumption_ablations}
\end{figure}

The reference configuration attains linear alignment
\(0.995\pm0.000\) and MCC \(0.994\pm0.000\)
(Fig.~\ref{fig:axis_assumption_ablations}). Violating constant conditional
entropy primarily affects the stronger axis-alignment criterion. With a
bias-free readout, increasing the across-class standard deviation of the
conditional-entropy term to \(2\) nats reduces MCC to
\(0.662\pm0.009\), while linear alignment remains
\(0.960\pm0.001\). Allowing the theoretically matched affine readout largely
absorbs the resulting class-dependent logit offsets, restoring linear
alignment to \(0.990\pm0.001\) and MCC to \(0.962\pm0.002\).

Moderate conditional dependence is comparatively well tolerated. At Gaussian
copula correlation \(0.5\), linear alignment and MCC remain \(0.971\) and
\(0.968\), respectively. Mean-matched Gamma and Weibull conditionals likewise
retain MCC \(0.990\) and \(0.984\), whereas the matched log-normal condition
reduces MCC to \(0.668\). These results indicate that exponentiality and
conditional independence are useful sufficient conditions but that modest
departures from either assumption need not eliminate axis alignment in this
finite-sample setting.

Signed latent support produces a sharper degradation: MCC falls to \(0.770\)
when latent signs are flipped with probability \(0.1\), and to \(0.049\) at
probability \(0.5\). By contrast, LeakyReLU and even a linear output attain MCC
\(0.987\) and \(0.968\), respectively, under the regularized reference
configuration. Removing weight decay leaves linear alignment near \(0.999\)
but reduces MCC to \(0.829\pm0.029\), compared with MCC \(0.994\) at weight
decay \(0.03\).

These ablations should not be interpreted as tests of whether each assumption
is individually necessary. The assumptions of \cref{thm:axis} jointly provide
a sufficient population-level guarantee, whereas the experiments measure the
solutions selected by a particular finite-sample optimization procedure, which is a more practical scenario.
Accordingly, persistence of axis alignment after an assumption is relaxed
shows empirical robustness in this setup.
Conversely, degradation identifies practically important boundaries.

\subsection{Non-identifiability in the Undercomplete Regime ($m<n'$)}
\label{app:exp:non_id}

\begin{figure}[t]
    \centering
    \includegraphics[width=1\linewidth]{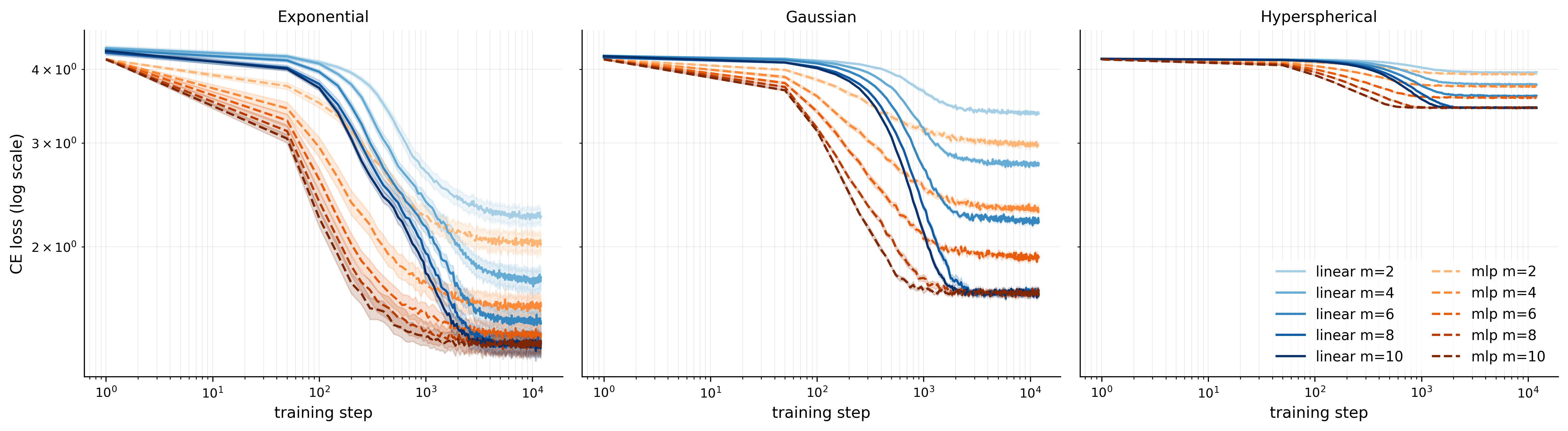}
    \caption{\textbf{Training Dynamics under undercomplete regime.}
    CE loss as a function of gradient update step for linear (solid blue)
    and MLP (dashed orange) encoders, across three latent distributions with encoder dimensions
    $m \in \{2, 4, 6, 8, 10\}$ (lighter shades indicate smaller $m$).
    Shaded areas indicate standard error of the mean over 10 random seeds.
    }
    \label{fig:nonid_dynamics}
\end{figure}

For the \emph{undercomplete} experiments we fixed the latent and effective dimensions at
$n = n' = 8$ and the number of classes at $|\mathcal{Y}| = 64$.
The teacher weight matrix $W^{*} \in \mathbb{R}^{|\mathcal{Y}| \times n}$ was constructed via
the low-rank factorisation $W^{*} = U^{*}A^{*}$, where $U^{*}$ and $A^{*}$ were
built from the leading columns of independent QR decompositions of random Gaussian
matrices with flat (unit) singular values, so that $\operatorname{rank}(W^{*}) = n'$
by construction.
Soft labels $p^{*}(y \mid z) = \operatorname{softmax}(W^{*}z)$ were computed from
latents drawn from three distributions: exponential $\operatorname{Exp}(1)$,
standard Gaussian, and uniform hyperspherical.

Two encoder architectures sharing the same input/output dimensions were compared: a
bias-free linear encoder $h(z) = Az$ and an MLP with three hidden layers of width
128, ReLU activations, and a linear output projection to $\mathbb{R}^{m}$ with no
final nonlinearity.
In both cases a bias-free linear readout $U \in \mathbb{R}^{|\mathcal{Y}| \times m}$ was
trained jointly with the encoder under soft cross-entropy.
The encoder dimension $m$ was swept over $\{2, 4, 6, 8, 10\}$, straddling the theoretical threshold at $m = n' = 8$.
Optimisation used Adam ($\beta = (0.9, 0.999)$, $\varepsilon = 10^{-8}$) with
learning rate $5 \times 10^{-4}$, batch size $512$, and $12{,}000$ gradient steps
without weight decay.
After training, $\operatorname{KL}(p^{*}\|q)$ was evaluated by
sampling with $N = 10{,}000$.
Every configuration was repeated over ten random seeds ($42 + i$,
$i \in \{0,\ldots,9\}$), with model initialisations re-seeded deterministically as
a function of all hyperparameters to ensure full reproducibility.
\cref{fig:nonid} reports the mean $\pm$ s.e.m.\ across seeds.

We also report the training dynamics in \cref{fig:nonid_dynamics}. Through three latent distributions (Exponential, Gaussian and Hyperspherical), the MLP consistently attains lower final CE than the linear encoder at every $m < n' = 8$, with the gap closing as $m$ approaches the capacity threshold; at $m \geq n'$, all architectures converge to comparable values, consistent with the \cref{thm:nonlinear} prediction.


\subsection{Compressed Sensing Recovery}
\label{app:cs}

For the compressed sensing experiments we set the full latent dimension to $n = 16$, the effective dimension to $n' = 8$, and the number of classes to $|\mathcal{Y}| = 64$.
The measurement matrix $A^{*} \in \mathbb{R}^{n' \times n}$ was drawn as a random
Gaussian matrix with $\ell_2$-normalised columns. The readout $U^{*} \in \mathbb{R}^{|\mathcal{Y}| \times n'}$ was drawn from
an independent Gaussian, and the teacher was set
to $W^{*} = U^{*}A^{*}$ with $\operatorname{rank}(W^{*}) = n'$ by construction.
Latents $z$ were $k$-sparse non-negative: the support of size $k$ was drawn uniformly at
random and the nonzero entries sampled from $\operatorname{Exp}(1)$, with all other
entries set to zero.
The sparsity level was swept over $k \in \{1, 2, 3\}$. Since \cref{thm:cs} requires
$2k \leq n'$, all three values satisfy the \emph{NSP} condition.

A bias-free linear encoder $h(z) = Az$ with $A \in \mathbb{R}^{m \times n}$, $m = n'=8$,
was trained jointly with a bias-free linear readout $U \in \mathbb{R}^{|\mathcal{Y}| \times m}$
under soft cross-entropy against teacher labels $p^{*}(y \mid z) = \operatorname{softmax}(W^{*}z)$.
Optimisation used Adam ($\beta = (0.9, 0.999)$, $\varepsilon = 10^{-8}$) with initial
learning rate $10^{-3}$ and a cosine annealing schedule over $20{,}000$ gradient
steps at batch size $512$ without weight decay.
Throughout training we logged $\KL(p^{*}\|q)$ and the row-space alignment
$\cos(\theta_{\max}) = \sigma_{\min}(Q_{A}^{\top}Q_{A^{*}})$,
where $Q_{A}$ and $Q_{A^{*}}$ are orthonormal bases of $\operatorname{row}(A)$ and
$\operatorname{row}(A^{*})$ obtained via QR decomposition to measure the similarity between the $\mathcal{N}(A)$ and $\mathcal{N}(A^*)$.

Three recovery methods were applied to the compressed activations $Az$:
supervised Lasso ($\ell_1$ regression with $\alpha = 5 \times 10^{-3}$ and
non-negativity constraint, solved per sample), dense linear OLS (fitted on
5{,}000 training samples and applied to 1{,}000 test samples), and unsupervised
dictionary learning followed by sparse coding (sklearn \texttt{DictionaryLearning}
with $n$ components, $\alpha = 5 \times 10^{-3}$, coordinate-descent fitting,
and trained on $h(z_{\mathrm{train}}) = Az_{\mathrm{train}}$
without access to $z$).
Recovery quality was assessed by $R^{2}$ for supervised methods and by the mean
correlation coefficient (MCC) with Hungarian matching between recovered and
ground-truth latent dimensions for all methods.
Each configuration was repeated over ten independent seeds
($42 + i$, $i \in \{0, \ldots, 9\}$).
Results are reported in \cref{fig:cs}.

\subsection{Real-data Analyses}\label{app:exp:real}
\textit{Vision models.}  
We examined four ImageNet-trained networks: VGG-19 BN, ResNet-50, ViT-b/16, and DINOv2 ViT-S/14.  
All images from the 50 000-sample ImageNet-1K validation set were resized to $224\times224$ pixels and normalised according to the respective preprocessing pipelines.  
We extracted activations from layer \texttt{fc6} (4096 units) in VGG-19, the global-average-pooled penultimate layer (2048 units) in ResNet-50, and the class token (768 and 384 units, respectively) in the two Vision Transformers.

\textit{Large-language models.}  
For GPT-2 small (124 M parameters), Pythia-70 M, and TinyLlama-1.1 B we generated single-token sentences containing each of the 100 000 most frequent Wikipedia unigrams.  
We recorded the last-token MLP activations from the final transformer block and computed semantic similarity using GPT-2-small embeddings (768-d), thereby avoiding circularity in the MIS calculation.

\textit{Biological data.}  
Neuronal firing rates were taken from the inferior-temporal cortex recordings of \citet{majaj2015simple}.  
Following the original preprocessing (square-root transformation and per-neuron normalisation), we analysed 58 neurons from subject 0 and 110 neurons from subject 1, each viewing 3 200 natural images.

\textit{Evaluation metrics.}  
Neuron interpretability was quantified with the Mechanistic Interpretability Score (MIS) of \citet{zimmermann_measuring_2024}, using 100 reference items and 10 queries per reference.  
To assess representational overlap between two models we computed the Maximum Matching Unit (MMU): for every neuron in model 1 we took the largest absolute Pearson correlation with any neuron in model 2.  
Correlations reported in Fig.~\ref{fig:real_data}C–E were tested two-sided against the null hypothesis $\rho=0$.



\end{document}